\documentclass[10pt,journal,compsoc]{IEEEtran}
\usepackage{amsmath}
\usepackage{tabularx}
\usepackage{booktabs}
\usepackage{graphicx}

\newcolumntype{L}[1]{>{\raggedright\arraybackslash}m{#1}}
\newcolumntype{C}[1]{>{\centering\arraybackslash}p{#1}}

\usepackage{longtable}
\usepackage{supertabular} 
\usepackage{multirow}
\usepackage[colorlinks=true,linkcolor=green,urlcolor=green,citecolor=green]{hyperref}

\ifCLASSOPTIONcompsoc
  \usepackage[nocompress]{cite}
\else
  \usepackage{cite}
\fi
\ifCLASSINFOpdf
\else
\fi
\begin{document}
%
\title{Recent Advances in Embedding Methods for Multi-Object Tracking: A Survey}
%
%
%
%

\author{Gaoang~Wang,~\IEEEmembership{Member,~IEEE,}
        Shengyu~Hao,
        Mingli~Song,~\IEEEmembership{Senior Member,~IEEE,}
        and~Jenq-Neng~Hwang,~\IEEEmembership{Fellow,~IEEE}
\IEEEcompsocitemizethanks{
\IEEEcompsocthanksitem G. Wang and S. Hao are with the Zhejiang University-University of Illinois at Urbana-Champaign Institute, Zhejiang University, China.\\
E-mail: gaoangwang@intl.zju.edu.cn, shengyuhao@zju.edu.cn
\IEEEcompsocthanksitem M. Song is with the College of Computer Science and Technology, Zhejiang University, China.\\
E-mail: brooksong@zju.edu.cn
\IEEEcompsocthanksitem J.-N. Hwang is with the Department of Electrical and Computer Engineering, the University of Washington, USA.\\
E-mail: hwang@uw.edu}
}

\IEEEtitleabstractindextext{%
\begin{abstract}
Multi-object tracking (MOT) aims to associate target objects across video frames in order to obtain entire moving trajectories. With the advancement of deep neural networks and the increasing demand for intelligent video analysis, MOT has gained significantly increased interest in the computer vision community. Embedding methods play an essential role in object location estimation and temporal identity association in MOT. Like other computer vision tasks, such as image classification, object detection, re-identification, and segmentation, embedding methods in MOT have large variations. However, they have never been systematically analyzed and summarized.
In this survey, we first conduct a comprehensive overview with an in-depth analysis of embedding methods in MOT from seven different perspectives, including patch-level embedding, single-frame embedding, cross-frame joint embedding, correlation embedding, sequential embedding, tracklet embedding, and cross-track relational embedding.
Then, we analyze the advantages of existing state-of-the-art methods according to their embedding strategies on widely used MOT benchmarks. Finally, some critical yet under-investigated areas and future research directions are discussed.
\end{abstract}

}

\maketitle

\IEEEdisplaynontitleabstractindextext

%
\IEEEpeerreviewmaketitle

\IEEEraisesectionheading{\section{Introduction}\label{sec:introduction}}

%
%
%
%
\IEEEPARstart{M}{ulti-object} tracking (MOT) has been widely studied in recent years, aiming to associate detected objects across video frames to obtain the entire moving trajectories.
With the development of tracking algorithms, MOT can be applied in many tasks, such as traffic flow analysis \cite{tang2018single,tang2019cityflow,wang2019anomaly,hsu2020traffic}, human behavior prediction and pose estimation \cite{wu2019unsupervised,gu2019efficient,gu2019multi,jalal2019multi}, autonomous driving assistance \cite{chaabane2021deft,hu2019joint}, and even underwater animal abundance estimation \cite{wang2016closed,chuang2016underwater,dawkins2017open}.

The flow of MOT systems can be mainly divided into two parts, \textit{i.e}., the embedding model and the association algorithm. With multiple successive frames as input, the object locations and track identities (IDs) are estimated via the embedding techniques and association approaches. 
MOT is challenging due to the presence of illumination changes, occlusions, complex environments, fast camera movement, unreliable detections, and varying low-image resolutions \cite{dendorfer2021motchallenge}. In addition, the tracking performance can be affected by individual steps of tracking algorithms, such as detection, feature extraction, affinity estimation, and association. These result in significant variations and uncertainty. With the recent progress in representation learning with deep neural networks, embedding methods play an essential role in object location estimation and temporal identity association in MOT. 
In this survey, we focus on the review of embedding learning rather than association methods \cite{ho2020two,zhang2022bytetrack,zhang2022robust,zou2022compensation}, though association methods are also important in MOT.

Like other computer vision tasks \cite{summaira2021recent,chen2014ranking,jiang2019learning,li2011graph}, such as image classification, object detection, object retrieval, crowd counting, visual tracking, re-identification (Re-ID), and segmentation, embedding methods in MOT also have large variations. However, the embedding learning approaches in MOT have never been systematically analyzed and summarized. 
Some embedding methods combine multi-task heads \cite{yang2021online,zhang2021fairmot,yu2022relationtrack,lu2020retinatrack,wang2020towards}, including box regression, object classification, and re-identification. Some embedding methods consider spatial-temporal correlations \cite{voigtlaender2019mots,zhou2020tracking,pang2020tubetk,sun2020simultaneous,wan2021multiple}, combining both appearance and motion information. Some methods exploit the interaction relationships among objects with correlation and attention to learn the track embeddings \cite{wang2021track,meinhardt2022trackformer,zeng2022motr}. The large deviation of embedding methods motivates us to conduct a comprehensive survey from an embedding perspective and discuss several under-investigated embedding areas and future directions.

\begin{figure*}[!t]
\begin{center}
\includegraphics[width=0.99\linewidth]{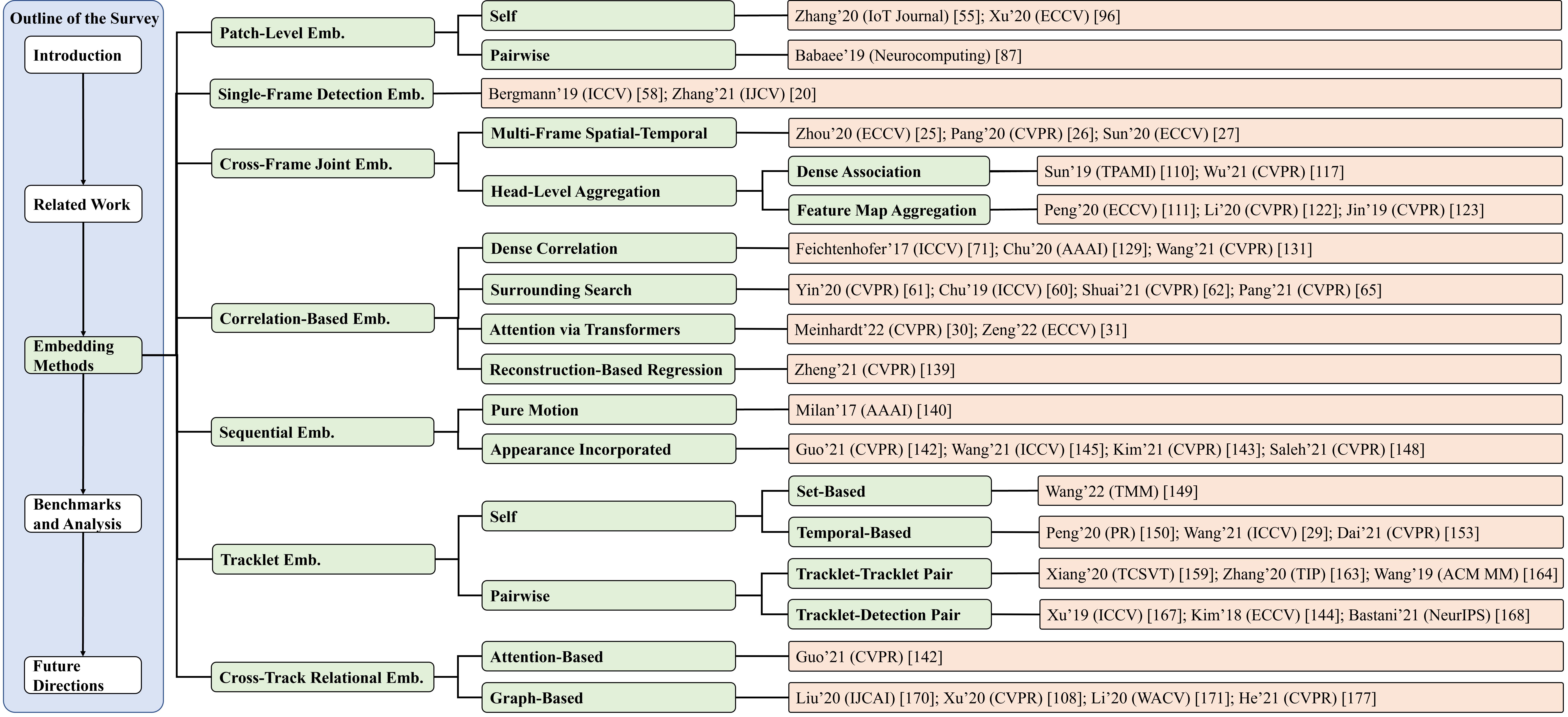}
\end{center}
\vspace{-10pt}
   \caption{A taxonomy of MOT embedding methods. The green and red boxes show the categories and representative references, respectively.}
\label{fig:framework}
\vspace{-10pt}
\end{figure*}

There are some existing surveys \cite{xu2019deep,emami2020machine,ciaparrone2020deep,park2021multiple,luo2021multiple} of MOT published in recent years. Specifically, Xu et al. \cite{xu2019deep} summarize some deep learning-based trackers and deep neural network structures. Emami et al. \cite{emami2020machine} focus on the review of model-based multiple hypotheses tracking with machine learning techniques in detection, filtering, and association. Ciaparrone et al. \cite{ciaparrone2020deep} provide a review of how deep learning is adopted in MOT, including detection, feature extraction, affinity computation, and association. Park et al. \cite{park2021multiple} review the evolution of MOT in recent decades, focusing on deep learning techniques and investigating the recent advances in MOT. Luo et al. \cite{luo2021multiple} provide a review of the MOT systems and discuss approaches from different aspects. Unlike all existing surveys, we focus on embedding learning in MOT, \textit{i.e}., how to learn object-level representative features for the MOT task, and comprehensively analyze state-of-the-art methods according to the embedding strategies. The main contributions of this survey are summarized as follows: 

1) We categorize and summarize existing embedding methods designed for MOT and make an in-depth and comprehensive analysis by discussing the advantages and limitations of the methods. The summary provides insights for future algorithm design and new topic exploration. 
2) We summarize the widely used benchmarks and analyze the state-of-the-art approaches according to the embedding methods. 
3) We attempt to discuss several important research directions with under-investigated issues related to embedding techniques and take a step towards future trends in MOT.

The outline of this survey is summarized as follows. We first demonstrate related works, including the most relevant tasks of MOT, in Section \ref{sec:related_work}. A taxonomy and detailed review of embedding methods are provided in Section \ref{sec:emb_method}. We then summarize the evaluation metrics and analyze the state-of-the-art approaches according to the embedding methods in Section \ref{sec:data_eval}. We discuss several under-investigated issues and point out the trends and potential future studies in Section \ref{sec:trend}. Conclusions are drawn in Section \ref{sec:conclude}.

\section{MOT Related Tasks}
\label{sec:related_work}

Three tasks, \textit{i.e}., single object tracking (SOT), video object detection (VOD), and re-identification (Re-ID), are highly related to MOT. Many embedding methods in the MOT field are inspired by these tasks. A comparison and brief review of these tasks are demonstrated in this section.

\subsection{Single Object Tracking}

Single-object tracking (SOT), also named visual object tracking (VOT), aims to estimate an unknown visual target trajectory when only an initial state of the target is available \cite{marvasti2021deep}. The tracking target is purely determined by the first frame and does not rely on any categories. Inspired by deep learning breakthroughs \cite{krizhevsky2012imagenet,simonyan2014very,he2016deep} in ImageNet large-scale visual recognition competition (ILSVRC) \cite{russakovsky2015imagenet} and visual object tracking (VOT) challenges \cite{kristan2018sixth,kristan2019seventh}, embedding learning methods have attracted considerable interest in the visual tracking community. Many works attempt the tracking by learning discriminative target representations, such as learning distractor-aware \cite{zhu2018distractor} or target-aware \cite{li2019target} features; leveraging different types of deep features such as context information \cite{morimitsu2018multiple,gao2019graph} or temporal features \cite{zhu2018end,chen2019multi}; exploring of low-level spatial features \cite{cen2018fully,fan2019siamese}; employing correlation-guided attention modules to exploit the relationship between the template and RoI feature maps \cite{du2020correlation}; and computing correlations between attentional features \cite{yu2020deformable}.

Unlike SOT, the initial states of objects are unknown in the MOT task, requiring pre-defined categories for tracking. As a result, MOT methods either adopt off-the-shelf detection following a tracking-by-detection scheme \cite{karthik2020simple,zhang2020multiplex,baisa2021occlusion,yang2021remot} or exploit the joint detection and tracking models \cite{bergmann2019tracking,voigtlaender2019mots,zhou2020tracking,yang2021online,zhang2021fairmot}. Some works \cite{feng2019multi,chu2019famnet,yin2020unified,shuai2021siammot,liang2022one,lin2021global,pang2021quasi} incorporate embedding methods from SOT into MOT with correlation and attention techniques.

\subsection{Video Object Detection}

Video object detection (VOD) aims to detect objects across multiple video frames by jointly recognizing objects and estimating locations \cite{jiao2021new}. 
Embedding methods are also important in the VOD task. Embedding learning can filter the features of a video frame, select relatively representative features, propagate key features for detection, and delineate the key areas that deserve feature filtering in subsequent frames.
For example, Guo et al. \cite{guo2019progressive} propose a special attention network known as the progressive sparse local attention framework to propagate features between different video frames. Xiao and Lee \cite{xiao2018video} use a module that calculates similar correlations to locally align the feature map. Chai \cite{chai2019patchwork} introduces Patchwork, a method that uses the attention mechanism to predict the position of an object in the next frame to solve the video object detection problem. Bertasius et al. \cite{bertasius2018object} propose a spatial-temporal sampling network that introduces deformable convolution to detect video frames.

Some embedding methods in VOD can be directly applied to MOT \cite{feichtenhofer2017detect}. However, unlike VOD, MOT usually does not require the recognition of objects unless objects from multiple categories are tracked jointly. Besides, track ID prediction is essential in the MOT task but not required in VOD. As a result, the embedding methods in MOT should be discriminative among objects even in the same category.

\subsection{Re-Identification}

Re-identification (Re-ID) aims at verifying object identity from different collections of images, usually from varying angles, illumination, and poses in different cameras \cite{ye2021deep,chen2017person,zheng2016person}. 
With the advancement of deep neural networks and increasing demand for intelligent video surveillance, Re-ID has significantly increased interest in the computer vision community. Much progress has been made in feature representation learning in Re-ID. Such embedding methods include global feature learning that extracts global feature representation for each image without additional annotation cues \cite{zheng2017person}; local feature learning that aggregates part-level local features to formulate a combined representation \cite{zhao2017deeply,yao2019deep,sun2018beyond}; auxiliary feature learning that improves the feature representation learning using auxiliary information like attributes \cite{su2016deep,lin2019improving,matsukawa2016person}; and video feature learning that learns video representation for video-based Re-ID using multiple image frames and temporal information \cite{liu2015spatio,dai2018video}.

Several MOT works \cite{feng2019multi,ye2020cost,baisa2021occlusion} employ Re-ID-based strategies to learn object appearance embeddings.
Unlike the Re-ID task that does not require temporal information between query and gallery images, temporal consistency is a more important cue in MOT.

\section{A taxonomy of MOT Embedding Methods}

\label{sec:emb_method}

Embedding methods are essential for object location estimation and ID association. The definition of embedding can be described as follows. Given the input data $\mathcal{X}$ and an embedding model $f_{\boldsymbol{w}}(\cdot)$ parameterized by $\boldsymbol{w}$, we aim to obtain object-level representations $\boldsymbol{z}$ from
\begin{equation}
    \boldsymbol{z}=f_{\boldsymbol{w}}(\mathcal{X}).
\label{eq:define_emb}
\end{equation}
For simplicity, we use $f(\cdot)$ to represent $f_{\boldsymbol{w}}(\cdot)$ in the following sections, without loss of generality. The input data $\mathcal{X}$ can include a combination of sequential image frames $\boldsymbol{X}$, cropped detection images $\boldsymbol{D}$, tracklets $\mathcal{T}$, 
and other types of data or modalities. The embedding models $f(\cdot)$ focus on deep learning-based methods. Other hand-crafted operations, such as color histograms, SIFT, HOG, and LBP, are not covered in the paper.

Our proposed taxonomy of MOT embedding methods is shown in Fig.~\ref{fig:framework}. The representative methods shown in the figure are based on the impact, \textit{i.e.}, either received sufficient citations or very recent progress published in top journals and conferences. 
We categorize the commonly used MOT embedding methods into seven groups, including patch-level embedding, single-frame embedding, cross-frame joint embedding, correlation-based embedding, sequential embedding, tracklet embedding, and cross-track relational embedding. 
The embedding methods are largely dependent on the type of input data. Some methods can take advantage of temporal information in the model design, like cross-frame joint embedding, sequential embedding, and tracklet embedding. Some methods can utilize the spatial information of the entire frames, like single-frame detection embedding and cross-frame joint embedding. Some only adopt object-level Re-ID models for feature extraction, like patch-level embedding. 
Different embedding strategies may not be mutually exclusive. One method may combine multiple embedding strategies to boost the tracking performance.
For each category, we introduce the representative algorithms and then discuss the weaknesses and strengths.


\subsection{Patch-Level Box Image Embedding}

Patch-level box image embedding attempts to extract object-level features from cropped detection images, which is essential for the tracking-by-detection scheme.
Some approaches treat MOT as a Re-ID problem. Most of the existing methods \cite{karthik2020simple,zhang2020multiplex,baisa2021occlusion,yang2021remot} in this category focus on embedding individual detections, while a few methods \cite{baisa2019online,yoon2018online,babaee2019dual} try to model the relation of two detected objects directly with pairwise embedding strategies. As a result, we categorize the patch-level box image embedding into two subgroups, \textit{i.e}., self-embedding and pairwise embedding.

\subsubsection{Self-Embedding}

Given the cropped box image based on the detection $\boldsymbol{D}_i$, the self-embedding represents the appearance feature of the detected object. This can be formulated as follows, 
\begin{equation}
    \boldsymbol{z}_i = f(\boldsymbol{D}_i),
\label{eq:crop_emb}
\end{equation}
where $f(\cdot)$ represents the embedding network, $\boldsymbol{z}_i$ represents the embedding of detection $\boldsymbol{D}_i$. 

The Re-ID framework with siamese networks is widely used for embedding to learn discriminative features to distinguish different object identities.
Some existing approaches \cite{karthik2020simple,zhang2020multiplex,baisa2021occlusion,chen2017enhancing,feng2019multi} treat the same objects across different frames as individual classes and employ cross-entropy loss for ID classification to learn the embedding of the cropped detection images. Such methods follow the very conventional embedding strategy that extracts object features by classification.
Some methods \cite{yang2021remot,chen2018real,ristani2018features,shen2018tracklet,li2021semi} adopt triplet loss and softmax-based contrastive loss to learn discriminative embeddings within batch samples, in which detections from the same object are treated as positive samples, and those from different objects are treated as negative samples. For the triplet loss, tuples of anchor, positive, and negative samples are constructed and exploited for distance metric learning, with the same strategy used in FaceNet \cite{schroff2015facenet}. Compared with cross-entropy loss, triplet loss can better distinguish the subtle difference among different IDs, especially for objects with similar color and pose. For the softmax-based contrastive loss, maximizing the similarity of positive pairs \cite{khosla2020supervised} is adopted. Unlike triplet loss which only takes three samples each time, contrastive loss proposed in \cite{li2021semi} compares multiple relations among different pairs from tracklets in the batch and shows promising performance in the embedding. 


Most detection-based embedding methods are based on cropped boxes. Except for bounding boxes, some works take advantage of segmentation masks for learning embeddings. Segmentation-based MOT can improve the effectiveness of the embedding, especially for partially occluded objects, and alleviate the influence of the background regions in the bounding box. For example, based on the instance segmentation results from \cite{neven2019instance}, PointTrack
\cite{xu2020segment}
treats the pixels of the cropped image as 2D point clouds, and the points from the foreground target and background environment are jointly learned for each cropped detection image.

There are several benefits of the patch-level self-embedding method. First, it is more flexible with variant detection algorithms, Re-ID models, and association methods, since each part can be trained separately. As a result, it can always combine off-the-shelf state-of-the-art detectors and Re-ID models. Second, since it takes cropped images as individual samples, it can employ a large amount of data in training. Compared with fully annotated video data, large-scale cropped image datasets are easier to collect. On the other side, the weakness is also obvious. It takes multiple stages in training, resulting in sub-optimal solutions. Besides, temporal information is hard to be learned in the embedding with cropped image samples.

\subsubsection{Pairwise Embedding}

\begin{figure}[!t]
\begin{center}
\includegraphics[width=0.7\linewidth]{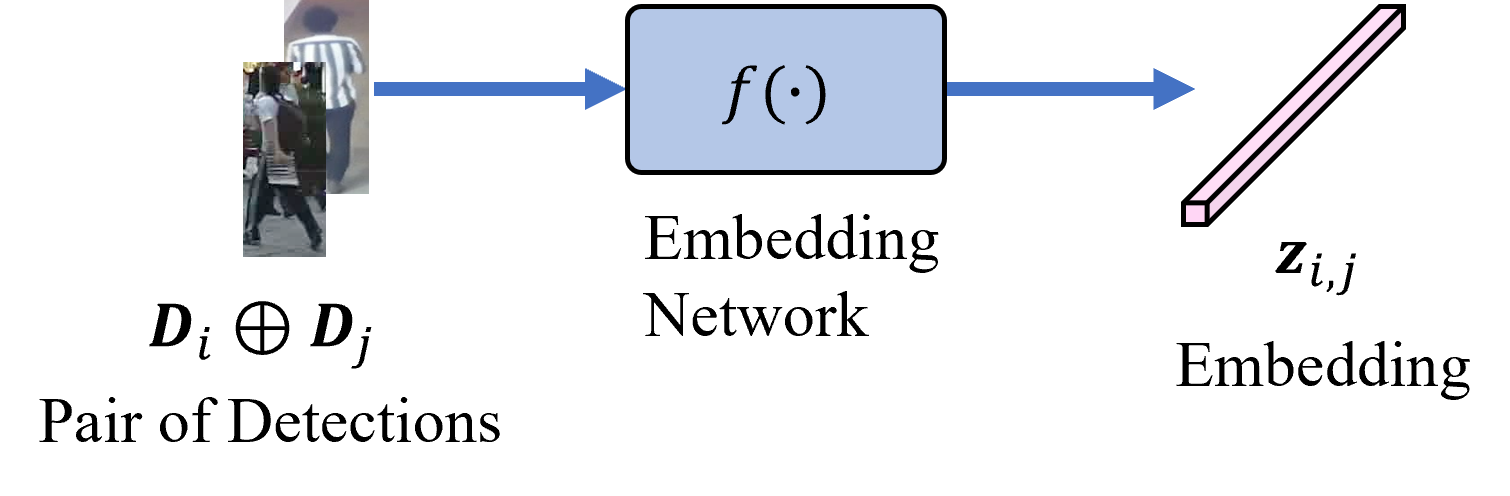}
\end{center}
\vspace{-10pt}
   \caption{Pairwise patch-level embedding. $\bigoplus$ represents the concatenation.}
\label{fig:pair_det_emb}
\vspace{-10pt}
\end{figure}

Rather than learning embeddings from individual detections, pairwise embedding networks take pairs of detections and directly learn the similarity between two detected objects, as shown in Fig.~\ref{fig:pair_det_emb}. A binary classifier is usually adopted to indicate whether the two detections belong to the same target, formulated as follows,
\begin{equation}
    \boldsymbol{z}_{i,j} = f(\boldsymbol{D}_i,\boldsymbol{D}_j),
\label{eq:jnt_crop_emb}
\end{equation}
where $\boldsymbol{z}_{i,j}$ is the pairwise embedding of two detections $\boldsymbol{D}_i$ and $\boldsymbol{D}_j$.

Some works \cite{baisa2019online,yoon2018online,babaee2019dual} follow a similar embedding framework in which concatenations of detection pairs are regarded as input. Then binary cross-entropy loss is adopted to distinguish whether two detections belong to the same object. 
Instead of concatenating input images, some works concatenate intermediate feature maps for embedding. 
For example, OTCD \cite{liu2019real}, a real-time online tracker in the compressed domain, follows this strategy, in which an appearance convolutional neural network is designed to take a pair of object features from detection to ensure a reliable association.
In addition, some works also combine motion cues into embedding. Specifically, the SiameseCNN tracker \cite{leal2016learning} concatenates both cropped image pairs and the corresponding optical flow maps as input, followed by a siamese convolutional neural network to learn descriptors encoding local spatial-temporal structures between the two input image patches and aggregate pixel values and optical flow information.
FANTrack \cite{baser2019fantrack} is proposed to take advantage of both appearance and motion features of paired objects with two-stage learning, where the first stage learns a similarity function that combines visual and 3D bounding box data to yield robust matching costs, and the second stage trains a CNN to predict discrete target assignments from the computed pairwise similarities.
For pairwise embedding, it has similar strengths to patch-level self-embedding. With paired input, the discrimination is easier to be learned among different IDs. 

\subsection{Single-Frame Detection Embedding}

\begin{figure}[!t]
\begin{center}
\includegraphics[width=0.7\linewidth]{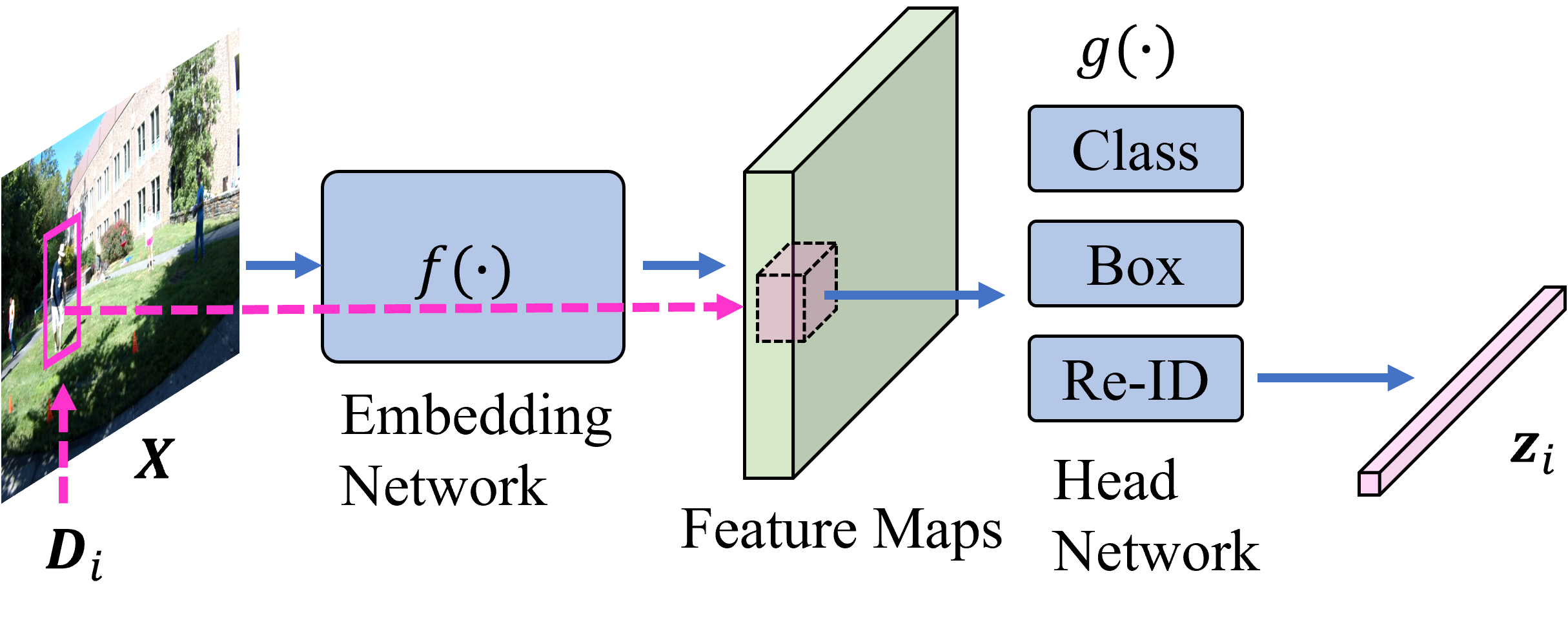}
\end{center}
\vspace{-10pt}
  \caption{Single-frame joint detection embedding.}
\label{fig:frame_emb}
\vspace{-10pt}
\end{figure}

Frame-based embedding jointly learns detection and Re-ID features in an end-to-end manner, as shown in Fig.~\ref{fig:frame_emb}. Given an input frame $\boldsymbol{X}$, the network learns the discriminative feature of each detection as follows,
\begin{equation}
    \{\boldsymbol{z}_i|i =\{1,2,...,|\mathcal{D}|\}\} = f(\boldsymbol{X}),
\label{eq:fr_emb}
\end{equation}
where $|\mathcal{D}|$ is the number of detections of the input frame, $\boldsymbol{z}_i$ is the embedding for the corresponding detection $\boldsymbol{D}_i$.

For single-frame based embedding, existing works usually follow joint detection and tracking framework and learn object-level embedding from detector-based architectures. Similar to object detection, such methods are designed to estimate object classes and locations. Furthermore, they are also required to ensure that the learned object embeddings are object-sensitive, \textit{i.e}., embeddings from different objects should keep dissimilar even if they are from the same class label. 
Some works use anchor-based frameworks for embedding learning, which are built on top of state-of-the-art detectors.
Specifically, some \cite{bergmann2019tracking,yang2021online} adopt Faster R-CNN \cite{ren2015faster}; RetinaTrack \cite{lu2020retinatrack} follows RetinaNet \cite{lin2017focal}; Wang et al. \cite{wang2020towards} exploit the YOLOv3 framework with DarkNet-53 \cite{redmon2018yolov3}.
However, as mentioned in \cite{zhang2021fairmot}, anchors are not fit for Re-ID features in nature. As a result, some works \cite{liu2022online,zhang2021fairmot} adopt anchor-free frameworks, which are built upon CenterNet \cite{zhou2019objects} with deep layer aggregation (DLA) \cite{yu2018deep,zhou2020tracking,yin2021center} used in the backbone architecture.

One of the major challenges for jointly learning embeddings for detection and Re-ID arises from the conflict between the two tasks. The detection task aims to identify object categories, such as pedestrians and vehicles, from the background, while Re-ID embeddings aim to distinguish distinct objects rather than the class. Some works decouple the embeddings from different tasks to address the issue with multi-task learning.
JDE \cite{wang2020towards} is one of the pioneer works that learn Re-ID embeddings built on top of detectors. The feature pyramid network (FPN) is employed in JDE, and task-specific heads are added on multiple FPN scales. 
Similarly, RetinaTrack \cite{lu2020retinatrack} designs task-specific post-FPN layers for classification and bounding box regression and generates embedding vectors simultaneously.
FairMOT \cite{zhang2021fairmot} is one of the representative methods that address the competition of multiple tasks in joint detection and tracking, which adopts the anchor-free object detection architecture CenterNet \cite{zhou2019objects}. In FairMOT, a single image frame is fed to an encoder-decoder network to extract high-resolution feature maps, followed by two homogeneous branches for detecting objects and extracting re-ID features, respectively. Then the features at the predicted object centers are used for tracking. Authors of FairMOT find that point-based embeddings perform much better than anchor-based embeddings in the tracking. 
Following the multi-task learning framework of FairMOT, UTrack \cite{liu2022online} designs an unsupervised Re-ID framework based on strong and weak supervision of the association cues from two frames.
RelationTrack \cite{yu2022relationtrack} is another method that attempts to alleviate the contradiction between detection and Re-ID tasks. Authors devise a global context disentangling module that decouples the learned representation into detection-specific and Re-ID-specific embeddings. They further develop a guided Transformer encoder module by combining the powerful reasoning ability of the Transformer encoder and deformable attention to consider the global semantic relation in the tracking.
Moreover, CCPNet \cite{Xu_2021_ICCV} decouples different tasks with multiple decoders to learn instance segmentation and ID-based embeddings for multi-object tracking and segmentation (MOTS) tasks. Meanwhile, a novel data augmentation strategy named continuous copy-paste is presented to fully exploit the pixel-wise annotations provided by MOTS to actively increase the number of instances as well as unique instance IDs in training.
Yang et al. \cite{yang2021online} adopt a similar embedding strategy for decoupling. In addition, they expand the training data by using the historical positions of the target combined with the prediction of the motion
model to improve the temporal correlation of extracted appearance features and employ the differentiable multi-objective tracking metric, \textit{i.e}., dMOTA and dMOTP \cite{xu2020train}, in training. 
MTrack \cite{yu2022towards} presents a novel embedding strategy apart from the detection head, namely multi-view trajectory contrastive learning, in which the discriminative representations are learned based on the contrast between a dynamically updated memory bank of trajectories and object features.


Compared with patch-level embedding, single-frame embedding methods do not rely on pre-trained detectors and do not require extra computation and storage to crop the detection patches. This one-stage training strategy increases the embedding efficiency. However, frame-based methods cannot efficiently utilize crop-based Re-ID datasets for training, causing a degradation for generalization. Moreover, temporal consistency is still not well-exploited in the embedding framework.

\subsection{Cross-Frame Joint Embedding}

To jointly learn the appearance and temporal features across multiple frames, cross-frame embedding plays an important role in MOT. The embedding can be formulated as follows,
\begin{equation}
    \{\boldsymbol{z}_i^t|i =\{1,2,...,|\mathcal{D}^t|\}\} = f([\boldsymbol{X}^{t-\tau};\boldsymbol{X}^{t-\tau+1};...;\boldsymbol{X}^t]),
\label{eq:cf_emb}
\end{equation}
where $[\boldsymbol{X}^{t-\tau};\boldsymbol{X}^{t-\tau+1};...;\boldsymbol{X}^t]$ represents the concatenation of multiple frames from $t-\tau$ to $t$. Some approaches \cite{zhou2020tracking,pang2020tubetk,sun2020simultaneous} adopt embedding networks, such as 3D convolution and convolutional LSTMs, that learn spatial-temporal feature maps for tracking, and some approaches \cite{sun2019deep,peng2020chained,xu2021transcenter} extract features for individual frames and then aggregate embeddings to model temporal relations in the task-specific heads. These two sub-categories are demonstrated in the following sub-sections.

\subsubsection{Multi-Frame Spatial-Temporal Embedding}

\begin{figure}[!t]
\begin{center}
\includegraphics[width=0.75\linewidth]{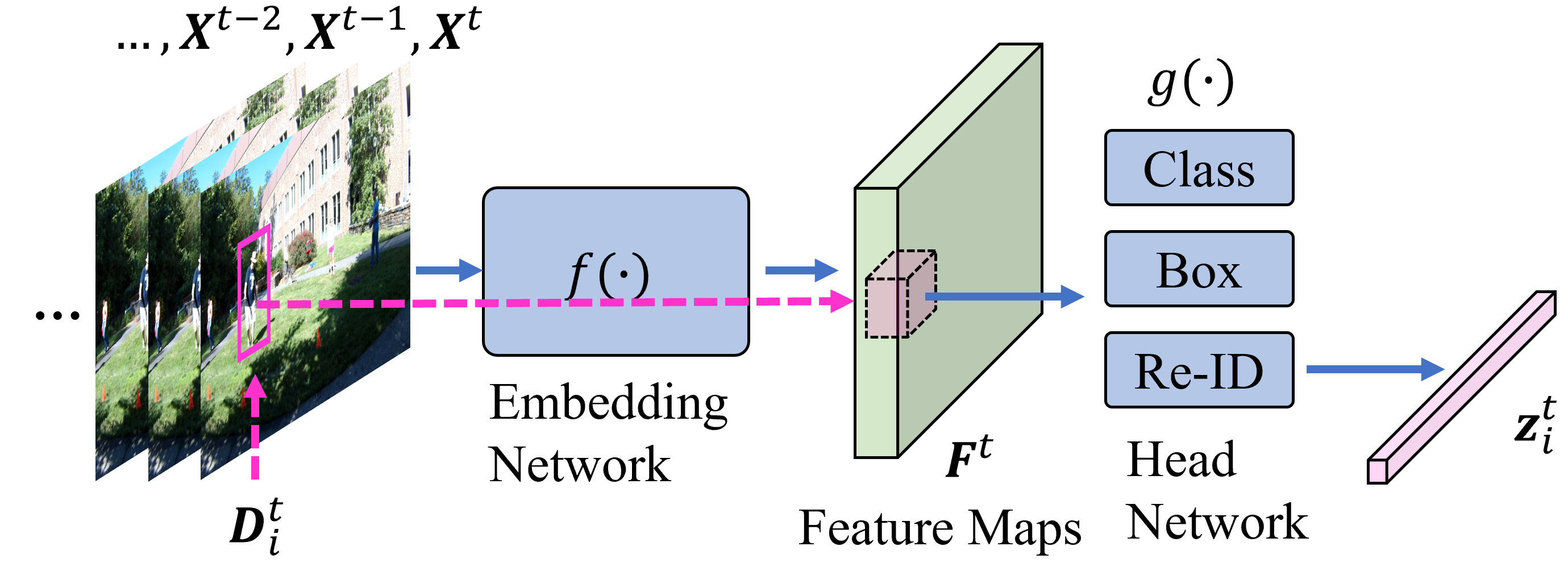}
\end{center}
\vspace{-10pt}
   \caption{Multi-frame spatial-temporal embedding.}
\label{fig:multi_frame_emb}
\vspace{-10pt}
\end{figure}

The multi-frame spatial-temporal embedding can be formulated as follows,
\begin{equation}
\begin{aligned}
& \boldsymbol{F}^t = f([\boldsymbol{X}^{t-\tau};\boldsymbol{X}^{t-\tau+1};...;\boldsymbol{X}^t]), \\
& \{\boldsymbol{z}_i^t|i =\{1,2,...,|\mathcal{D}^t|\}\} = g(\boldsymbol{F}^t),
\label{eq:st_emb}
\end{aligned}
\end{equation}
where $\boldsymbol{F}^t$ is the intermediate feature maps obtained from the backbone embedding network $f(\cdot)$ that learns spatial-temporal information from sequential frames $[\boldsymbol{X}^{t-\tau};\boldsymbol{X}^{t-\tau+1};...;\boldsymbol{X}^t]$, and $g(\cdot)$ is a head network that generates the final embeddings $\{\boldsymbol{z}_i^t|i =\{1,2,...,|\mathcal{D}^t|\}\}$ of objects. This subcategory can be regarded as an early fusion of multiple frames for embedding, as shown in Fig.~\ref{fig:multi_frame_emb}.

TubeTK \cite{pang2020tubetk} is a representative method that takes advantage of spatial-temporal information from short video clips. 3D-ResNet \cite{hara2018can,he2016deep} is employed as the backbone to generate feature maps, followed by the regression of bounding tubes in a 3D manner, trained with a combination of multiple losses, including GIOU \cite{rezatofighi2019generalized}, focal loss \cite{lin2017focal}, and binary cross-entropy loss. 
Similarly, DMM-Net \cite{sun2020simultaneous} employs 3D convolution to learn spatial-temporal embeddings given multiple frames to generate tubes and predicts multi-frame motion, classes, and visibility to generate tracklets. 
Wan et al. \cite{wan2021multiple} take multiple frames using an encoder-decoder architecture with temporal priors embedding based on short connections to estimate multi-channel trajectory maps simultaneously, including the presence map, appearance map, and motion map.
Rather than tracking objects with bounding boxes, Track R-CNN \cite{voigtlaender2019mots} is the first work that extends the MOT task to multi-object tracking and segmentation (MOTS), conducting tracking and instance segmentation simultaneously. The MOTS follows the framework of Mask R-CNN \cite{he2017mask} and aggregates the feature maps from multiple frames using 3D convolution and convolutional LSTM \cite{xingjian2015convolutional} for temporal context embedding, which is an end-to-end method of segmentation-based embedding for prediction and association. 
Some works also combine the prediction from previous frames to guide the embedding and association. For example, CenterTrack \cite{zhou2020tracking} employs the CenterNet framework \cite{zhou2019objects} and concatenates a pair of sequential frames and the heat map of the previous frame for joint embedding, object center location estimation, as well as the size and offset prediction.

With the capability of learning temporal consistency with 3D neural networks and LSTMs, motion features can be incorporated into the embedding framework. 
However, current spatial-temporal embedding usually only considers a few frames for joint embedding. As a result, the learned temporal motion features are not robust enough to model the diverse movement of objects. Learning the long-time dependency also needs to be further studied in future works.

\subsubsection{Head-Level Feature Aggregated Embedding}

\begin{figure}[!t]
\begin{center}
\includegraphics[width=0.85\linewidth]{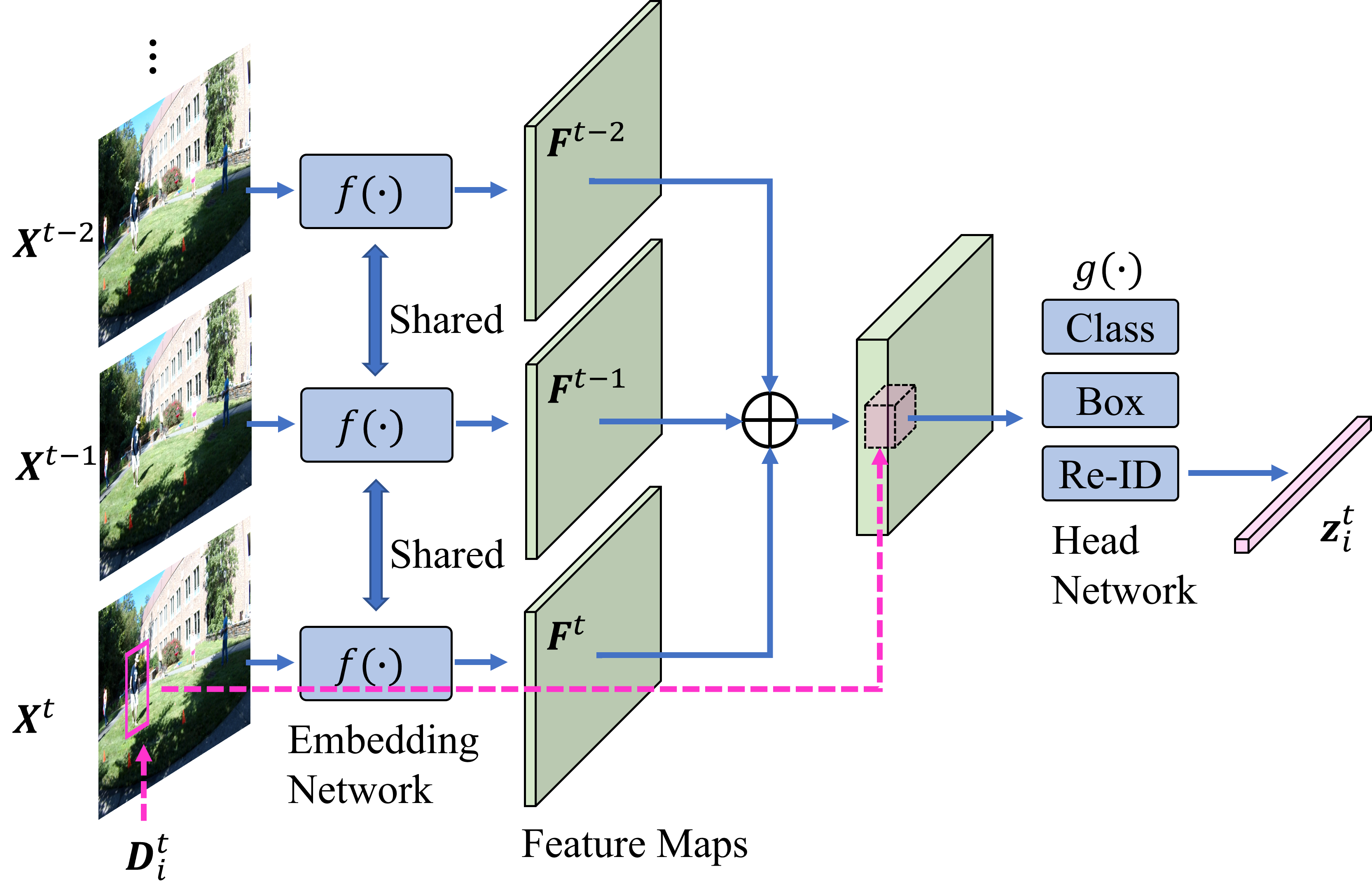}
\end{center}
\vspace{-10pt}
   \caption{Head-level aggregated embedding.}
\label{fig:multi_frame_emb2}
\vspace{-10pt}
\end{figure}

Slightly different from the multi-frame spatial-temporal embedding, methods of this sub-category usually extract features of input frames independently and then aggregate the feature maps in the head network for embedding, which can be formulated as follows, 
\begin{equation}
\begin{aligned}
& \boldsymbol{F}^t = f(\boldsymbol{X}^{t}), \\
& \{\boldsymbol{z}_i^t|i =\{1,2,...,|\mathcal{D}^t|\}\} = g([\boldsymbol{F}^{t-\tau};\boldsymbol{F}^{t-\tau+1};...;\boldsymbol{F}^t]),
\label{eq:st_emb}
\end{aligned}
\end{equation}
where $f(\cdot)$ and $g(\cdot)$ represents the backbone network and head network, respectively. This sub-category can be regarded as the late fusion of spatial-temporal embeddings, as shown in Fig.~\ref{fig:multi_frame_emb2}.

\textbf{Dense pair aggregation}. 
Some works aggregate embeddings for dense pairwise matching. Specifically, DAN \cite{sun2019deep} presents a deep affinity network that predicts the dense affinity of detection center locations between a pair of frames using the features extracted from different layers of VGGNet \cite{simonyan2014very}, leading to reliable online tracking performance.
Similar to DAN \cite{sun2019deep}, DEFT \cite{chaabane2021deft} designs a matching head to aggregate the embeddings from pairs of frames. Besides that, motion forecasting with LSTM is exploited in the matching head for association in DEFT. 
TraDeS \cite{wu2021track} also follows a dense pair aggregation scheme, where the proposed cost volume-based association module extracts point-wise re-ID embedding features by the backbone to construct a cost volume that stores matching similarities between the embedding pairs in adjacent two frames. Then the motion-guided feature warper module is employed to jointly learn the 2D and 3D locations and masks based on the propagation of offsets with motion cues using deformable convolution \cite{dai2017deformable}.
The dense pair aggregation can be applied in the graph neural networks (GNN). For example, GSDT \cite{wang2021joint} aggregates node embeddings of detected objects based on GNN using GraphConv \cite{morris2019weisfeiler} given a pair of sequential frames.
The dense pair aggregation can also be employed in the pose embedding and tracking scenarios. For example, POINet \cite{ruan2019poinet} proposes an end-to-end pose-guided insight network for the data association in multi-person pose tracking, which jointly learns feature extraction, similarity estimation, and identity assignment. POINet includes two main components, \textit{i.e}., a pose-guided representation network and an ovonic insight network, where the pose-guided representation network is used for feature extraction, and the ovonic insight network aggregates the dense object pairs from two frames and estimates the similarity based on forward and backward insight.

\textbf{Feature map aggregation}. 
Apart from aggregating dense embeddings, some works concatenate feature maps from a pair of frames for prediction. 
Chained-tracker \cite{peng2020chained} is one of the representative embedding approaches, which adopts the stacked feature maps for bounding box regression and ID verification following Faster R-CNN \cite{ren2015faster}, providing a straightforward and standard framework for feature map aggregation methods. 
Except for the prediction from concatenated feature maps, Li et al. \cite{Li_2020_CVPR} also construct spatial-temporal optimization with both motion and geometry cues from sequential stereo images in training. 
A similar idea can be utilized in pose-tracking scenarios. Jin et al. \cite{Jin_2019_CVPR} propose a unified pose estimation and tracking framework with two main components, \textit{i.e}., SpatialNet, and TemporalNet. The produced heatmaps of sequential frames from the SpatialNet are aggregated to learn the keypoint embedding and temporal instance embedding in the TemporalNet module for tracking. 
HandLer \cite{huang2022forward} is a hand tracker which takes features from two consecutive frames and designs a forward propagation with flow estimation and temporal feature aggregation, followed by a backward regression for offset prediction and detection.
Unicorn \cite{yan2022towards} presents a unified method that can simultaneously solve four tracking problems (SOT, MOT, VOS, MOTS) with a single network, which is the first work that accomplishes the great unification of the tracking network architecture and learning paradigm. Based on the feature maps generated from two consecutive frames, a feature interaction module is exploited to build pixel-wise correspondence between two frames. Based on the correspondence, a target prior is generated by propagating the reference target to the current frame. Then the target prior and the visual features are fused and sent to the detection head to get the tracked objects for all tasks.
Several recent works adopt Transformer architecture for aggregating feature maps for embedding.
Time3D \cite{li2022time3d} extends the tracking to 3D for autonomous driving scenarios and introduces a spatial-temporal information flow module that aggregates geometric and appearance features to predict robust similarity scores across all objects in current and past frames based on the attention mechanism of the Transformer.
UTT \cite{ma2022unified} presents a unified Transformer tracker that combines MOT and SOT in one paradigm. After extracting frame features, the information is aggregated by a track Transformer, which consists of one target decoder to extract target features, one proposal decoder to produce candidate search areas, and one target transformer to predict the target localization.
P3AFormer \cite{zhao2022tracking} employs a Transformer-based model that tracks objects as pixel-wise distributions. Specifically, consecutive frame features are fed into a Transformer pixel decoder to obtain the pixel-wise representations. Then the temporal contextual information is extracted based on pixel-wise feature propagation, followed by an object decoder that takes masks, image features, and queries for detection and embedding.


Compared with multi-frame spatial-temporal embedding, head-level aggregated embedding methods encode each frame individually with shared backbones.
However, it lacks low-level pixel-wise correlation features between frames for detection and association.

\subsection{Correlation-Based Embedding}

\begin{figure}[!t]
\begin{center}
\includegraphics[width=0.8\linewidth]{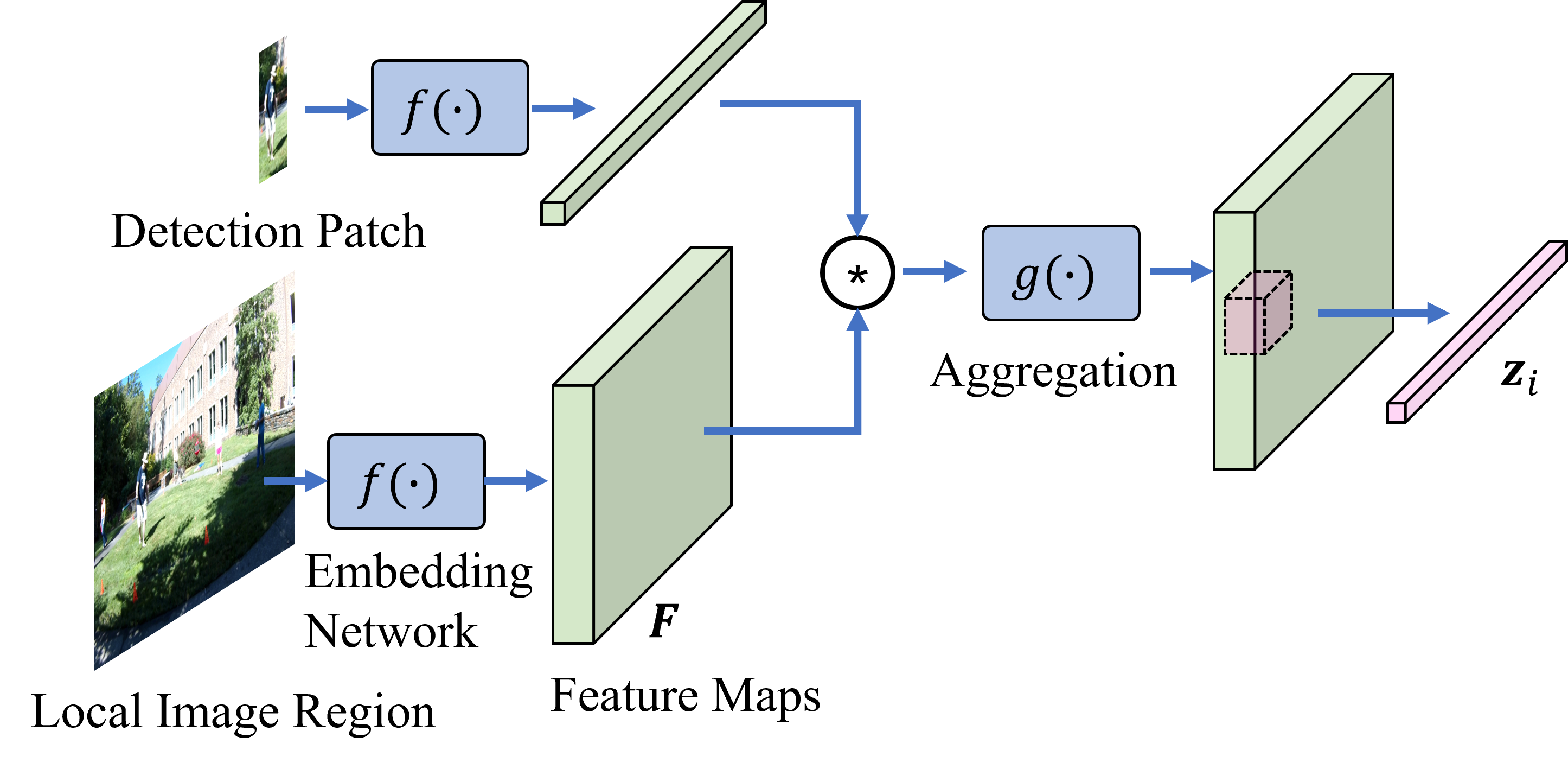}
\end{center}
\vspace{-10pt}
   \caption{Correlation-based embedding.}
\label{fig:corr_emb}
\vspace{-10pt}
\end{figure}

Inspired by the SOT methods, target location can be refined via the correlation between detections and generated feature maps, as shown in Fig.~\ref{fig:corr_emb}. The correlation-based approaches can be represented as follows,
\begin{equation}
    \boldsymbol{z}_i = g(\boldsymbol{F} \ast f(\boldsymbol{D}_{i})),
\label{eq:cor_emb}
\end{equation}
where $\ast$ is denoted as the correlation operation, $g(\cdot)$ and $f(\cdot)$ are denoted as the embedding networks. In correlation-based embedding, features from the previous frame are usually regarded as correlation kernels of the next frame. As a result, more accurate object locations can be obtained based on the correlation.

\textbf{Dense correlation}. 
Some works estimate the dense correlation feature maps. 
For example, D\&T \cite{feichtenhofer2017detect} is one of the pioneer works based on correlation-based embedding. After obtaining the feature maps from convolutional networks, the convolutional cross-correlation between the feature responses of adjacent frames is computed to estimate the local displacement at different feature scales. Meanwhile, the object class and box are estimated in a separate detection branch. 
DASOT \cite{chu2020dasot} uses a similar idea as D\&T to compute correlation feature maps. In their implementation, the correlation operation is only conducted in a local square neighborhood since displacements of the same targets across two adjacent frames are limited. In addition, DASOT integrates data association and SOT in a unified framework, in which dense correlation feature maps are estimated for the temporal association, built upon truncated ResNet-50 with the feature pyramid network (FPN) \cite{lin2017feature}.
Rather than only considering the temporal correlation between a pair of frames, STLC \cite{wang2021multiple} estimates both multi-scale spatial correlation and temporal correlation with dense feature maps. Specifically, the spatial correlation module is established to model the topological relationship between targets and their surrounding environment, and the temporal correlation module is proposed to establish frame-to-frame matches over convolutional feature maps in the different layers to align and propagate temporal context.

\textbf{Surrounding search}. 
Some works conduct correlations between individual detections and surrounding local regions via SOT algorithms. 
For example, following deep tracker SiamFC \cite{bertinetto2016fully}, UMA \cite{yin2020unified} uses the siamese network to calculate the correlation between anchor samples with both positive and negative local regions. Built upon lightweight AlexNet \cite{krizhevsky2012imagenet}, the network adopts triplet loss for discriminative learning to generate embeddings.
FAMNet \cite{chu2019famnet} formulates an end-to-end model for MOT with a combination of feature extraction, affinity estimation, and multi-dimensional assignment. The SOT is incorporated the proposed affinity network, where the cross-correlation is conducted between the anchor feature and local regions in adjacent frames for each anchor object in the center frame. 
VAN \cite{leeACCV2020van} presents a versatile affinity network that can perform the entire MOT process in a single network, including target-specific SOT to handle incomplete detection issues, affinity computation between target and candidates, and decision of tracking termination.
SiamMOT \cite{shuai2021siammot} conducts the motion modeling exploration by leveraging a region-based Siamese MOT network. It combines a region-based detection network with two motion models, \textit{i.e}.,  an implicit motion model (IMM) and an explicit motion model (EMM), inspired by the literature on siamese-based SOT. In the proposed EMM, the cross-channel correlation is estimated between the detected patch and local regions of the next frame, following the Faster R-CNN detector.
OMC \cite{liang2022one} follows the joint detection and tracking framework. After ID embedding, a re-check network is proposed to conduct cross-correlation between previous embeddings and current feature maps. The re-check network innovatively expands the role of ID embedding from data association to motion forecasting by effectively propagating previous tracklets to the current frame.
QDTrack \cite{pang2021quasi} models the correlation with contrastive learning, which extracts contrastive embeddings around temporal neighbors sampled from ROIs for both positive and negative instances, following the Faster R-CNN framework.
UAVMOT \cite{liu2022multi} presents a novel framework for tracking unmanned aerial vehicles. An ID feature update module is proposed in UAVMOT to calculate the correlation between the current frame and the top-k points from the previous feature map, which addresses the irregular motion and view change in 3D.

\textbf{Attention via Transformers}. 
With recent advances in visual Transformers, some approaches \cite{sun2020transtrack,meinhardt2022trackformer,zeng2022motr,chen2022patchtrack} adopt Transformer in MOT since transformers use pairwise attention that can fuse global information in the embedding and boost the tracking performance. The query-key mechanism plays a role as a correlation in tracking. The prediction can be obtained with multi-head attention \cite{vaswani2017attention} that measures the correlation between feature maps and track queries. 
For example, TransTrack \cite{sun2020transtrack} presents a baseline tracker with Transformer, which uses track queries and object queries in the Transformer decoder for both track prediction and object detection.
Similarly, Trackformer \cite{meinhardt2022trackformer} and MOTR \cite{zeng2022motr} also employ the Transformer decoder to estimate the correlation between previous tracks and current feature maps for prediction.
In Trackformer, the encoder-decoder computes attention between the input image features and the track as well as object queries and outputs bounding boxes with assigned IDs at each frame.
Built upon DETR \cite{carion2020end}, each track query of MOTR models the entire track of an object, which is transferred and updated frame-by-frame to perform iterative predictions in a seamless manner.

\textbf{Reconstruction-based regression}. 
Apart from approaches that use correlation for prediction, some works adopt reconstruction-based strategies with regression to learn the correlations for embedding. For example, built upon a variant of DLA-34 \cite{yu2018deep}, SOTMOT \cite{zheng2021improving} solves the ridge regression coefficient from the previous frame for each object and uses the learned coefficient to predict the label for the next frame. As a result, discriminative features can be learned around neighbors in local regions. 

Correlation-based embedding can learn the local-global correspondence. This type of embedding also can borrow state-of-the-art methods from SOT problems. When targets are partially occluded and lost from detectors, such trackers are still able to continually track the objects. However, drift becomes the major issue when long-time occlusion occurs, and trackers may gradually lose the targets.

\subsection{Sequential Embedding}

Another commonly used strategy to model temporal information in MOT is to use recurrent neural networks (RNNs) for sequential modeling. Such sequential embedding methods learn the dynamic update of transformations from the previous embedding to the current embedding. Embedding methods of this category can be formulated as follows,
\begin{equation}
    \boldsymbol{z}_i^t = f(\boldsymbol{z}_i^{t-1}, \boldsymbol{x}_i^{t}),
\label{eq:seq_emb}
\end{equation}
where $\boldsymbol{z}_i^{t-1}$ represents the historical embeddings, $\boldsymbol{x}_i^{t}$ represents the input to the current state.

\textbf{Pure motion}. 
Some works use sequential embedding to model motion features. They assume object trajectories usually have non-linear patterns, which can be modeled by variants of RNNs. 
Milan et al. \cite{milan2017online} present the first end-to-end model using RNN to predict object motion with deep learning. 
LM\_NN \cite{babaee2019dual} tests both gated recurrent unit (GRU) and long short-term memory (LSTM) for offset prediction of object location and size with the input of historical motions. 
Jiang et al. \cite{jiang2019graph} utilize historical motions to predict future motion with LSTM units and calculate aggregated affinity with both appearance and motion patterns using siamese networks. Besides, graph neural networks are adopted to obtain the association matrix by embedding the affinity.
Similarly, DEFT \cite{chaabane2021deft} adopts a motion forecasting module with LSTM. 

\textbf{Appearance incorporated}. 
Some works also take into consideration of appearance features in the sequential embedding.
TADAM \cite{guo2021online} presents a unified model with a synergy between position prediction and embedding association, where the two tasks are linked by temporal-aware target attention and distractor attention, as well as an identity-aware memory aggregation model. To better embed the temporal and contextual information in the track memory, the convolutional gated recurrent unit (ConvGRU) is employed in the identity-aware memory aggregation model to aggregate spatial-temporal appearance features.
Similarly, BLSTM-MTP \cite{kim2021discriminative} exploits bilinear LSTM \cite{kim2018multi} to encode spatial-temporal appearance information of the tracklet and trains a binary classifier on the positive and hard negative tracklets to learn discriminative embeddings.
Wang et al. \cite{wang2021general} present a multiple-node tracking (MNT) framework in which track trees are generated to link detections and tracks. After that, a recurrent tracking unit (RTU) is proposed to calculate scores of proposals in the built track tree, which takes the old appearance feature template, the appearance feature of the current node, the old hidden state, and the state feature of the current node as input for embedding.
Different from the previous methods that use sequential embedding for object-level features, PermaTrack \cite{tokmakov2021learning} augments the model with a spatial-temporal recurrent memory module, \textit{i.e.}, ConvGRU \cite{ballas2015delving}, to encode entire historical frames for feature map estimation, which is built on top of CenterTrack \cite{zhou2020tracking} and extends the architecture to an arbitrary length of the input.
ArTIST \cite{saleh2021probabilistic} designs a moving agent network, which is built upon a recurrent auto-encoder neural network that learns to reconstruct the tracklets of all moving agents with both appearance and motion cues.

Sequential embedding can learn the long dependency of the motion of targets. However, the standalone sequential embedding cannot achieve high performance. As a result, it is usually combined with other embedding methods, such as tracklet embedding, cross-track relational embedding, and single-frame detection embedding.

\subsection{Tracklet Embedding}

Following the tracking-by-detection paradigm, many approaches treat tracklets as individual units rather than detections, aiming at exploiting local and global temporal cues in the embedding. Typically, tracklet embedding can also be divided into two sub-categories: tracklet self-embedding and pairwise embedding.

\subsubsection{Self-Embedding}

Self-embedding can be formulated as follows,
\begin{equation}
    \boldsymbol{z}_i = f(\mathcal{T}_i),
\label{eq:track_emb1}
\end{equation}
where $\mathcal{T}_i=[\boldsymbol{D}_i^{t-\tau};\boldsymbol{D}_i^{t-\tau+1};...;\boldsymbol{D}_i^{t}]$ is the $i$-th tracklet, containing sequential detections of the same object from local intermediate prediction.

\textbf{Set-based}. 
Some works treat each tracklet as a disordered set. Then learning the tracklet embedding is similar to the set classification problem. 
For example, TBooster \cite{wang2022split} designs a tracklet refinement method with two proposed modules, \textit{i.e.}, splitter, and connector. Multi-head self-attention is exploited in the connector module to aggregate features of individual detections and learns the tracklet embedding with the combination of triplet loss and cross-entropy loss for the association.

\textbf{Temporal-based}. 
To better model the temporal cues, many works also take the temporal order into the embedding. 
For example, TPM \cite{peng2020tpm} presents a representative-selection network with bi-directional convolutional LSTM that encodes contextual and temporal information to select representative object features for tracklet embedding. 
Hsu et al. \cite{hsu2019multi} follow the temporal modeling of video-based person Re-ID framework \cite{gao2018revisiting}, which adopts temporal convolution to model temporal attention for feature aggregation to obtain tracklet embedding. 
Some works employ GNN for tracklet embedding. LGM \cite{wang2021track} is a reconstruct-to-embed framework that embeds the tracklet based on motion features with temporal convolution and GNN to learn the interaction among tracklets. 
LPC\_MOT \cite{dai2021learning} presents a novel proposal-based learnable framework, which models MOT as a proposal generation, proposal scoring, and trajectory inference paradigm on an affinity graph. Specifically, it concatenates both averaged appearance features, and the spatial-temporal features and then graph convolutional neural network (GCN) \cite{kipf2016semi} is further adopted for tracklet embedding.
PHALP \cite{rajasegaran2022tracking} aggregates the information from the previous tracklet for future appearance, location, and pose prediction in 3D space. Specifically, HMAR \cite{rajasegaran2021tracking} and HMMR \cite{kanazawa2019learning} with Transformer backbone are used in PHALP for appearance and pose embedding, respectively.
MeMOT \cite{cai2022memot} introduces a memory aggregator for tracklet embedding, which consists of three attention modules, including one short-term attention for noise removal, one long-term attention for long-range context encoding, and one fusion block for aggregation. 

Compared with other embedding methods, the tracklet self-embedding can incorporate motion information from nature. As a result, the appearance and motion features can be combined with a unified model. However, generating tracklets is tricky. The errors from the tracklet generation can cause degradation in the tracklet embedding.

\subsubsection{Pairwise Embedding}

Different from self-embedding, pairwise embedding jointly models the association relationship of the input tracklet pair. This kind of approach can be summarized as follows,
\begin{equation}
    \boldsymbol{z}_{i,j} = f(\mathcal{T}_i, \mathcal{T}_j),
\label{eq:track_emb2}
\end{equation}
where $\boldsymbol{z}_{i,j}$ is the pairwise embedding of tracklet $\mathcal{T}_i$ and $\mathcal{T}_j$.

\textbf{Tracklet-tracklet pair}. 
Many works model the pairwise embedding of two input tracklets. 
For example, CRF-RNN \cite{xiang2020end} learns unary and pairwise potentials based on the CRF model \cite{zheng2015conditional,larsson2017learning} using CNN and bidirectional LSTM architecture to encode the long-term dependencies between CRF nodes, where each graph node represents a pair of tracklets.
Following the multiple hypothesis tracking \cite{reid1979algorithm} framework, Zhang et al. \cite{zhang2020long} measure the input tracklets whether they belong to the same track using CNN and LSTM to model appearance and motion separately in the proposed appearance evaluation network and motion evaluation network, respectively.
TNT \cite{wang2019exploit} takes account of the temporal orders of the input tracklet pair based on the temporal connectivity for pairwise embedding, where the motion and appearance features are concatenated by temporal convolutions to learn the connectivity.

\textbf{Tracklet-detection pair}. 
Rather than tracklet-to-tracket pairs, some works model the joint embedding between tracklet and detection. In other words, $\mathcal{T}_j$ from Eq.~(\ref{eq:track_emb2}) may only contain a single detection. 
For example, TAMA \cite{yoon2021online} learns a pair of detection with softmax as binary classification in the joint-inference network \cite{leal2016learning}. Then a detection is compared with all previous detections within the tracklet, and LSTM is adopted to measure the similarity in the association network.
Inspired by the object relation module in \cite{hu2018relation}, STRN \cite{xu2019spatial} presents a spatial-temporal relation network, which is a unified framework for similarity measurement between a tracklet and an object. Various cues, such as appearance, topology, and location cues, are simultaneously encoded across time in STRN for calculating the final similarity measurement.
MHT-bLSTM \cite{kim2018multi} learns the embedding of tracklet-detection pair with the proposed bilinear LSTM, where the tracklet and detection are coupled in a multiplicative manner instead of the additive manner in conventional LSTM approaches. Such coupling resembles an online learned regressor at each time step, which is found to have Superior performance for appearance modeling.
Bastani et al. \cite{bastani2021self} propose a self-supervised MOT method with cross-input consistency, which takes two distinct inputs for the same sequence of video by hiding different information. Then a matching network is conducted to embed the concatenated features of the tracklet-detection pair for matching and association.
DMAN \cite{zhu2018online} adopts a spatial attention network to obtain the feature of each pair of the detection and a frame of the tracklet. Bi-directional LSTM (Bi-LSTM) is exploited to learn the temporal feature of the SAN in the temporal attention network. Then the final embedding is learned for each tracklet-detection pair.

Tracklet pairwise embedding has similar properties as tracklet self-embedding methods. Besides that, it can also learn the discriminative relations and temporal order among different tracklets. Since each pair of tracklets needs to be embedded separately, the computational cost is huge, which is the major weakness.

\subsection{Cross-Track Relational Embedding}

\begin{figure}[!t]
\begin{center}
\includegraphics[width=0.7\linewidth]{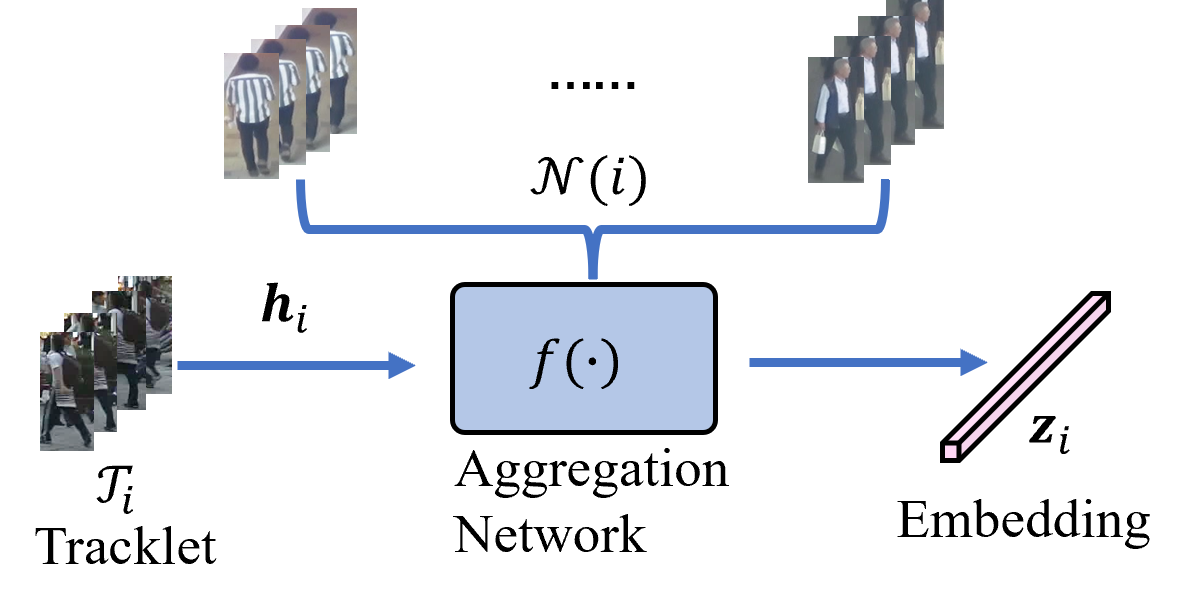}
\end{center}
\vspace{-10pt}
   \caption{Cross-track relational embedding.}
\label{fig:cross_emb}
\vspace{-10pt}
\end{figure}

Cross-track relational embedding aims at learning the object features based on the interactions with the neighboring tracks, as shown in Fig.~\ref{fig:cross_emb}. The embedding can be represented as follows,
\begin{equation}
    \boldsymbol{z}_{i} = f(\{\boldsymbol{h}_j|j \in i \cup \mathcal{N}(i)\}),
\label{eq:cross_emb}
\end{equation}
where $\boldsymbol{h}_j$ is the intermediate embedding of a detection or a tracklet; $\mathcal{N}(i)$ represents the connected neighbors of the $i$-th track defined in the approach.

\textbf{Learning relational embedding from attention}. 
The attention mechanism is one of the commonly used strategies to obtain track embeddings with the interactions of other tracks. 
For example, though TADAM \cite{guo2021online} is introduced in sequential embedding with ConvGRU to aggregate sequential features, it also considers cross-track relations in the embedding and presents the temporal-aware target attention and distractor attention for object feature extraction. 
Some Transformer-based trackers also exploit attention to obtain cross-track embeddings. 
For example, Trackformer \cite{meinhardt2022trackformer} and MOTR \cite{zeng2022motr} encode the track embeddings using the correlation with other tracks and objects via multi-head attention. 

\textbf{Learning relational embedding from graph}. 
Graph-based methods are also widely employed for cross-track embedding. 
For example, GSM \cite{liu2020gsm} presents a graph representation that takes both the feature of individual objects and the relations among objects into consideration. In GSM, local graphs are defined for each detected object and its k-nearest neighbors. Then the graph similarity model is designed for every two frames to measure the association via binary classification. 
DeepMOT \cite{xu2020train} proposes a deep Hungarian network that approximates the Hungarian matching algorithm and uses bi-directional RNN to model the association. In addition, differentiable dMOTA and dMOTP metrics are defined for end-to-end training. 
Li et al. \cite{li2020graph} design a near-online MOT method with an end-to-end graph network that can be dynamically updated. Specifically, it models the motion graph network and appearance graph network separately. The graph is built for detections in adjacent two frames. The final similarity is measured with a weighted sum of motion and appearance. 
Following the message passing network (MPN) \cite{gilmer2017neural,guo2018neural,battaglia2016interaction,battaglia2018relational}, Braso et al. \cite{braso2020learning} exploit the classical network flow formulation of MOT and define a fully differentiable framework, which aggregates neighborhood nodes for detection embedding. 
GMT \cite{he2021learnable} presents a learnable graph-matching method that models the relationships between tracklets and the intra-frame detections as a general undirected graph. Continuous quadratic programming is adopted for approximation to facilitate the end-to-end training of the GNN.
There are also many works that combine other embedding strategies and cross-track relational embedding in their methods.
For example, LGM \cite{wang2021track} not only employs tracklet embedding but also exploits the cross-track relations in the GCN with the proposed reconstruct-to-embed strategy to learn discriminative features from the information of other trackers. 
ArTIST \cite{saleh2021probabilistic} combines sequential embedding and cross-track relational embedding. Specifically, ArTIST presents a moving agent network (MA-Net), where MA-Net is a recurrent auto-encoder neural network that learns to reconstruct the tracklets of all moving agents potentially interacting with the tracklet of interest.
Based on the multi-frame spatial-temporal embedding, GSDT \cite{wang2021joint} also exploits node feature aggregation among different detections based on GNN given two sequential frames, which can better extract the relational features across targets. 
GTR \cite{zhou2022global} presents a global tracking Transformer that takes detections from sequential frames and calculates the cross-attention with trajectory queries.

Unlike other embedding strategies, cross-track embedding can learn the interaction among tracks. The discrimination between different IDs can be significantly exploited. It can also be combined with other embedding methods, like sequential and tracklet embedding, to enhance representation learning. However, a thorough analysis of the effectiveness of cross-tracking embedding has not been well studied, requiring more future research.

\section{Benchmarks and Analysis}
\label{sec:data_eval}

\subsection{Evaluation Metrics}

To evaluate the performance of MOT, CLEAR MOT \cite{bernardin2008evaluating}, ID-based metrics \cite{ristani2016performance}, and HOTA \cite{luiten2021hota} are the widely used measurements.

CLEAR MOT \cite{bernardin2008evaluating} measures the multiple object tracking accuracy (MOTA) and multiple object tracking precision (MOPT) between the detected boxes and ground truth boxes. The proposed MOTA and MOTP are based on the matched pairs, misses, false positives, and mismatches. However, MOTA overemphasizes detection performance rather than association. 

ID-based metrics \cite{ristani2016performance} compute the truth-to-result match that measures the tracking performance by how long the tracker correctly identifies targets. Specifically, a bipartite match associates one ground-truth trajectory to exactly one computed trajectory by minimizing the number of mismatched frames over all the available data-true and computed. Standard measures such as precision, recall, and F1-score are built on top of this truth-to-result match, namely IDP, IDR, and IDF1. However, IDF1 overemphasizes association performance rather than detection.

A higher order metric, namely higher order tracking accuracy (HOTA), is proposed in \cite{luiten2021hota} for evaluating MOT performance. HOTA explicitly balances the effect of performing accurate detection, association, and localization into a single unified metric for comparing trackers. HOTA decomposes into a family of sub-metrics that are able to evaluate each of five basic error types (detection recall, detection precision, association recall, association precision, and localization accuracy) separately, which enables clear analysis of tracking performance.

There are also other metrics for evaluation. For example, PR-MOTA is proposed in the UA-Detrac dataset \cite{wen2020ua} and takes account of the detection score along the PR-curve with the MOTA metric for MOT measurement. In the nuScenes dataset \cite{caesar2020nuscenes}, sAMOTA is adopted as the primary evaluation metric for the 3D MOT benchmark, in which an updated formulation of MOTA is used to adjust for the respective recall. 
Voigtlaender et al. \cite{voigtlaender2019mots} introduce the soft multi-object tracking and segmentation accuracy (sMOTSA) for MOTS-based tracking measurement.
Feng et al. \cite{feng2020samot} propose the SAIDF evaluation measure that focuses more on identity issues and fixes the insensibility and high computational cost problems of the previous measures, such as MOTA and IDF1. Furthermore, Valmadre et al. \cite{valmadre2021local} introduce local metrics, namely LIDF1 and ALTA, that are parametrized by a temporal horizon and thereby reveal the temporal ranges at which association errors occur. Such metrics provide more insights from different aspects of MOT evaluation.

\subsection{In-Depth Analysis on State-of-the-Art Methods}

\begin{table}[!t]
\caption{Summary of embedding methods on MOT17 benchmark.}
\centering
\vspace{-5pt}
\begin{tabular}{L{0.9cm}|L{0.7cm}|L{0.6cm}|L{1.0cm}|L{0.5cm}|L{0.6cm}L{0.4cm}L{0.6cm}}
\toprule
\textbf{Emb. Method} &  \textbf{Ref}. & \textbf{Year} & \textbf{Venue} & \textbf{Det}. & \textbf{MOTA} & \textbf{IDF1} & \textbf{HOTA} \\
\midrule

\multirow{11}{*}{\shortstack[l]{Patch}} & 
\cite{chen2017enhancing} & 2017 & CVPRW & Pub. & 50.0&51.3&41.3 \\
~ & \cite{shen2018tracklet} & 2018 & CoRR & Pub. & 51.5&46.9&38.5 \\
~ & \cite{chen2018real} & 2018 & ICME & Pub. & 50.9&52.7&41.2 \\
~ & \cite{yoon2018online} & 2018 & AVSS & Pub. & 48.3&51.1&40.3\\
~ & \cite{feng2019multi} & 2019 & ArXiv & Pub. & 54.7 & 62.3 & 47.1 \\
~ & \cite{liu2019real} & 2019 & Access & Pub. & 48.6&47.9&38.4 \\
~ &  \cite{karthik2020simple} & 2020 & ArXiv & Pub. & 61.7 & 58.1 & 46.9 \\
~ &  \cite{ye2020cost} & 2020 & ICRAI & Pub. & 49.7&51.5&39.0 \\
~ & \cite{baisa2021occlusion} & 2021 & J-VCIR & Pub. & 46.8&54.1&41.5\\
~ & \cite{yang2021remot} & 2021 & IVC & Priv. & 77.0&72.0&59.7\\
~ & \cite{li2021semi} & 2021 & ArXiv & Priv. & 73.3&73.2&59.8\\

\midrule
\multirow{5}{*}{\shortstack[l]{S-Fr}} &
\cite{bergmann2019tracking} & 2019 & ICCV & Pub. & 56.3 & 55.1 & 44.8 \\
~ &  \cite{yang2021online} & 2021 & AI & Pub. & 60.1 & 58.8 & 47.2 \\
~ & \cite{zhang2021fairmot} & 2021 & IJCV & Priv. & 73.7&72.3&59.3\\
~ & \cite{yu2022relationtrack} & 2022 & T-MM & Priv. & 73.8&74.7&61.0\\
~ & \cite{liu2022online} & 2022 & Neuroc. & Priv. & 73.5&70.2&58.7\\
~ & \cite{yu2022towards} & 2022 & CVPR & Priv. & 72.1&73.5&-\\

\midrule
\multirow{8}{*}{\shortstack[l]{X-Fr}} & 
\cite{sun2019deep} & 2019 & T-PAMI & Pub. & 52.4&49.5&39.3\\
~& \cite{zhou2020tracking} & 2020 & ECCV & Pub. & 61.5 & 59.6 & 48.2 \\
~ & \cite{pang2020tubetk} & 2020 & CVPR & Priv. & 63.0&58.6&48.0\\
~ & \cite{peng2020chained} & 2020 & ECCV & Priv. & 66.6&57.4&49.0\\
~ & \cite{zhou2020tracking} & 2020 & ECCV & Priv. & 67.8&64.7&52.2\\
~ & \cite{wang2021joint} & 2021 & ICRA & Priv. & 73.2&66.5&55.2\\
~ & \cite{xu2021transcenter} & 2021 & ArXiv & Priv. & 73.2&62.2&54.5 \\
~ & \cite{chen2022patchtrack} & 2022 & ArXiv & Priv. & 73.6&65.2&53.9\\
~ & \cite{yan2022towards} & 2022 & ECCV & Priv. & 77.2 &75.5&61.7\\
~ & \cite{zhao2022tracking} & 2022 & ECCV & Priv. & 81.2 &78.1&-\\

\midrule
\multirow{5}{*}{\shortstack[l]{Corr}} & 
\cite{chu2019famnet} & 2019 & ICCV & Pub. & 52.0 & 48.7 & - \\
~ & \cite{chu2020dasot} & 2020 & AAAI & Pub. & 49.5&51.8&41.5 \\
~ & \cite{leeACCV2020van} & 2020 & ACCV & Pub. & 57.4&57.9& - \\
~ & \cite{sun2020transtrack} & 2020 & ArXiv & Priv. & 75.2&63.5&54.1 \\
~ & \cite{pang2021quasi} & 2021 & CVPR  & Priv. & 68.7&66.3&53.9\\
~ & \cite{wang2021multiple} & 2021 & CVPR & Priv. & 76.5&73.6&60.7\\

\midrule
\multirow{6}{*}{\shortstack[l]{Seq}} &  
\cite{babaee2019dual} & 2019 & Neuroc. & Pub. & 45.1&43.2&-\\
~ & \cite{kim2021discriminative} & 2021 & CVPR & Pub. & 51.5&54.9&41.3 \\
~ &  \cite{girbau2021multiple} & 2021 & ArXiv & Pub. & 67.8 & 61.4 & 50.0 \\
~ & \cite{tokmakov2021learning} & 2021 & ICCV & Pub. & 73.1 & 67.2 & 54.2 \\
~ & \cite{tokmakov2021learning} & 2021 & ICCV & Priv. & 73.8&68.9&55.5\\
~ &\cite{wang2021general} & 2021 & ICCV & Priv. & 74.9&75.0&62.0\\

\midrule
\multirow{13}{*}{\shortstack[l]{Tracklet}}  &
\cite{babaee2018multiple} & 2018 & ArXiv & Pub. & 54.1 & 48.4 & 46.8 \\
~ & \cite{zhu2018online} & 2018 & ECCV & Pub. & 48.2&55.7&42.5 \\
~ & \cite{kim2018multi} & 2018 & ECCV & Pub. & 47.5&51.9&41.0\\
~ & \cite{xu2019spatial} & 2019 & ICCV  & Pub. & 50.9&56.0&42.6 \\
~ & \cite{wang2019exploit} & 2019 & MM & Pub. & 51.9&58.1&44.9 \\
~ &  \cite{zhang2020long} & 2020 & T-IP & Pub. & 54.9 & 63.1 & 48.4 \\
~ & \cite{peng2020tpm} & 2020 & PR & Pub. & 54.2 & 52.6 & 41.5 \\
~ & \cite{xiang2020end} & 2020 & T-CSVT & Pub. & 53.1 & 53.7 & 42.2 \\
~ & \cite{baisa2021robust} & 2021 & J-VCIR & Pub. & 45.4&39.9&34.0\\
~ & \cite{yoon2021online} & 2021 & IS & Pub. & 50.3&53.5&42.0 \\
~ & \cite{bastani2021self} & 2021 & NeurIPS & Pub. & 56.8 & 58.3 & 46.4 \\
~ & \cite{dai2021learning} & 2021 & CVPR & Pub. & 59.0 & 66.8 & 51.5 \\
~ & \cite{wang2022split} & 2022 & T-MM & Pub. & 61.5 & 63.3 & 50.5 \\
~ & \cite{cai2022memot} & 2022 & CVPR & Priv. & 72.5 & 69.0 & 56.9 \\

\midrule
\multirow{6}{*}{\shortstack[l]{X-Track}} & 
\cite{braso2020learning} & 2020 & CVPR & Pub. & 58.8 & 61.7 & 49.0 \\
~ & \cite{papakis2020gcnnmatch} & 2020 & ArXiv & Pub. & 57.3 & 56.3 & 45.4 \\
~ & \cite{liu2020gsm} & 2020 & IJCAI & Pub. & 56.4 & 57.8 & 45.7 \\
~ & \cite{xu2020train} & 2020 & CVPR & Pub. & 53.7 & 53.8 & 42.4 \\
~ & \cite{shan2020tracklets} & 2020 & ArXiv & Priv. & 76.2&68.0&57.9\\
~ & \cite{wang2021joint} & 2021 & ICRA & Priv. & 73.2&66.5&55.2\\
~ & \cite{zhou2022global} & 2022 & CVPR & Priv. & 75.3&71.5&59.1\\

\bottomrule
\end{tabular}
\vspace{-10pt}
\label{tab:SOTA_MOT17}
\end{table}

\begin{table}[!t]
\caption{Summary of embedding methods on MOT20 benchmark.}
\centering
\vspace{-5pt}
\begin{tabular}{L{0.9cm}|L{0.7cm}|L{0.6cm}|L{1.0cm}|L{0.5cm}|L{0.6cm}L{0.4cm}L{0.6cm}}
\toprule
\textbf{Emb. Method} & \textbf{Ref}. & \textbf{Year} & \textbf{Venue} & \textbf{Det}. & \textbf{MOTA} & \textbf{IDF1} & \textbf{HOTA} \\
\midrule

\multirow{4}{*}{\shortstack[l]{Patch}} &
\cite{karthik2020simple} & 2020 & ArXiv & Pub. & 53.6 & 50.6 & 41.7 \\
~ &\cite{baisa2021occlusion} & 2021 & J-VCIR & Pub. & 44.7 & 43.5 & 35.6 \\
~ & \cite{yang2021remot} & 2021 & IVC & Priv. & 77.4 & 73.1 & 61.2 \\
~ & \cite{li2021semi} & 2021 & ArXiv & Priv. & 65.2 & 70.1 & 55.3 \\

\midrule
\multirow{5}{*}{\shortstack[l]{S-Fr}}  &
\cite{bergmann2019tracking} & 2019 & ICCV & Pub. & 52.6 & 52.7 & 42.1 \\
~ & \cite{yang2021online} & 2021 & AI & Pub. & 59.3 & 59.1 & 47.1 \\
~ & \cite{zhang2021fairmot} & 2021 & IJCV & Priv. & 61.8 & 67.3 & 54.6 \\
~ & \cite{yu2022relationtrack} & 2022 & T-MM & Priv. & 67.2 & 70.5 & 56.5 \\
~ & \cite{liu2022online} & 2022 & Neuroc. & Priv. & 68.6 & 69.4 & 56.2 \\
~ & \cite{yu2022towards} & 2022 & CVPR & Priv. & 63.5 & 69.2 & - \\

\midrule
\multirow{2}{*}{\shortstack[l]{X-Fr}} &
\cite{xu2021transcenter} & 2021 & ArXiv & Pub. & 61.0 & 49.8 & 43.5 \\
~ & \cite{wang2021joint} & 2021 & ICRA & Priv. & 67.1 & 67.5 & 53.6 \\
~ & \cite{zhao2022tracking} & 2022 & ECCV & Priv. & 78.1 & 76.4 & - \\

\midrule
\multirow{1}{*}{\shortstack[l]{Corr}} & 
\cite{sun2020transtrack} & 2020 & ArXiv & Priv. & 65.0 & 59.4 & 48.9 \\

\midrule
\multirow{2}{*}{\shortstack[l]{Tracklet}} & 
\cite{dai2021learning} & 2021 & CVPR & Pub. & 56.3 & 62.5 & 49.0 \\
~ & \cite{wang2022split} & 2022 & T-MM & Pub. & 54.6 & 53.4 & 42.5 \\
~ & \cite{cai2022memot} & 2022 & CVPR & Priv. & 63.7 & 66.1 & 54.1 \\

\midrule
\multirow{2}{*}{\shortstack[l]{X-Track}} & 
\cite{braso2020learning} & 2020 & CVPR & Pub. & 57.6 & 59.1 & 46.8 \\
~ & \cite{papakis2020gcnnmatch} & 2020 & ArXiv & Pub. & 54.5 &	49.0 & 40.2 \\

\bottomrule
\end{tabular}
\vspace{-10pt}
\label{tab:SOTA_MOT20}
\end{table}

\begin{table}[!t]
\caption{Summary of embedding methods on KITTI MOT benchmark.}
\centering
\vspace{-5pt}
\begin{tabular}{L{0.9cm}|L{0.7cm}|L{0.6cm}|L{0.8cm}|L{0.6cm}|L{0.5cm}|L{0.6cm}L{0.6cm}}
\toprule
\textbf{Emb. Method} & \textbf{Ref}. & \textbf{Year} & \textbf{Venue} & \textbf{Mod}. & \textbf{Obj}. & \textbf{MOTA} & \textbf{HOTA} \\
\midrule

\multirow{5}*{Patch} 
& \cite{baser2019fantrack} & 2019 & ArXiv & V & C & 75.8&60.9\\
~& \cite{shenoi2020jrmot} & 2020 & IROS & V\&L & C & 85.1&69.6\\
~& \cite{mykheievskyi2020learning} & 2020 & ACCV & V & C & 87.8&68.5\\

~& \cite{mykheievskyi2020learning} & 2020 & ACCV & V & P & 68.0&50.9\\
~& \cite{shenoi2020jrmot} & 2020 & IROS & V\&L & P &45.3&34.2\\

\midrule
\multirow{1}*{S-Fr} 
&\cite{hu2019joint} & 2019 & ICCV & V & C & 84.3&73.2\\

\midrule
\multirow{8}*{X-Fr} & 
\cite{zhang2019robust} & 2019 & ICCV & V\&L & C & 83.2 & 62.1\\
~&\cite{wang2020pointtracknet} & 2020 & RAL & L & C & 67.6 & 57.2\\
~& \cite{zhou2020tracking} & 2020 & ECCV &V & C & 88.8 & 73.0\\
~& \cite{wu2021tracklet} & 2021 & IJCAI & L & C & 91.7 & 80.9 \\
~& \cite{chaabane2021deft} & 2021 & CVPR & V & C & 88.4 & 74.2 \\
~& \cite{wang2021ditnet} & 2021 & RAL & L & C & 84.5 & 72.2\\
~& \cite{huang2021joint} & 2021 & IROS & V\&L & C & 85.4 & 70.7\\
~& \cite{zhao2022tracking} & 2022 & ECCV & V & C & 91.2 & 78.4\\
~& \cite{zhou2020tracking} & 2020 & ECCV &V & P & 53.8 & 40.4\\
~& \cite{zhao2022tracking} & 2022 & ECCV & V & P & 67.7 & 49.0\\

\midrule
\multirow{5}*{Corr} & 
\cite{chu2019famnet} & 2019 & ICCV & V & C & 75.9 & 52.6 \\
~&\cite{hu2021monocular} & 2021 & ArXiv & V & C & 85.9 & 72.8\\
~&\cite{pang2021quasi} & 2021 & CVPR & V & C & 84.9 & 68.5\\

~& \cite{pang2021quasi} & 2021 & CVPR & V & P & 55.6 & 41.1\\
~&\cite{hu2021monocular} & 2021 & ArXiv & V & P & 51.8 & 41.1\\

\midrule
\multirow{3}*{Seq} 
&\cite{hu2019joint} & 2019 & ICCV & V & C & 84.3&73.2\\
~&\cite{tokmakov2021learning} & 2021 & ICCV & V & C & 91.3&78.0\\

~&\cite{tokmakov2021learning} & 2021 & ICCV & V & P & 66.0&48.6\\

\midrule
\multirow{1}*{Tracklet} 
&\cite{wang2021track} & 2021 & ICCV & V & C & 87.6&73.1\\

\midrule
\multirow{4}*{X-Track} 
&\cite{wang2021track} & 2021 & ICCV & V & C & 87.6&73.1\\
~&\cite{rangesh2021trackmpnn}& 2021 & ArXiv & V & C & 87.3&72.3\\

~& \cite{braso2020learning} & 2020 & CVPR & V & P & 46.2 & 45.3\\
~&\cite{rangesh2021trackmpnn}& 2021 & ArXiv & V & P & 52.1&39.4\\

\bottomrule
\end{tabular}
\label{tab:SOTA_KITTI}
\vspace{-10pt}
\end{table}

\begin{table*}[!t]
\caption{SOTA embedding methods on DIVOTrack. More details of the metrics can be found in \cite{luiten2021hota,bernardin2008evaluating,ristani2016performance}.}
\centering
\vspace{-5pt}
\begin{tabular}{L{1.2cm}|L{2.3cm}|L{1.0cm}L{1.0cm}L{1.0cm}L{1.0cm}L{1.0cm}L{1.0cm}L{1.0cm}L{1.0cm}L{1.0cm}}
\toprule
\textbf{Emb. Method} & \textbf{Tracker} & \textbf{HOTA}$\uparrow$ & \textbf{IDF1}$\uparrow$ & \textbf{MOTA}$\uparrow$ & \textbf{AssA}$\uparrow$ & \textbf{sMOTA}$\uparrow$ & \textbf{IDR}$\uparrow$ &\textbf{IDP}$\uparrow$ & \textbf{IDSW}$\downarrow$ & \textbf{Frag}$\downarrow$ \\
\midrule
Patch & ByteTrack \cite{zhang2022bytetrack} & 55.4 & 66.1 & 61.2 & 55.7 & 46.1 & 64.7 & 67.6 & 1,470 & 4,442\\
S-Fr & FairMOT \cite{zhang2021fairmot} & 54.0 & 64.4 & 59.4 & 55.1 & 45.2 & 60.8 & 68.4 & 1,283 & 5,183\\
X-Fr & CenterTrack \cite{zhou2020tracking} & 42.9 & 46.0 & 51.2 & 37.4 & 35.0 & 46.2 & 45.9 & 3,011 & 3,813\\
Corr & QDTrack \cite{pang2021quasi} & 46.8 & 52.6 & 53.4 & 43.6 & 38.3 & 51.1 & 54.1 & 2,300 & 4,288\\
Seq & PermaTrack \cite{tokmakov2021learning} & 48.2 & 53.8 & 56.2 & 44.3 & 38.8 & 55.7 & 52.1 & 2,183 & 3,109\\
Tracklet & LPC\_MOT \cite{dai2021learning} & 56.5 & 68.6 & 51.1 & 61.6 & 34.2 & 74.8 & 63.3 & 824 & 1,375\\
X-Track & GTR \cite{zhou2022global} & 46.8 & 62.9 & 60.0 & 51.8 & 44.0 & 64.3 & 61.7 & 1,450 & 3,982\\
\bottomrule
\end{tabular}
\vspace{-10pt}
\label{tab:SOTA_DIVO}
\end{table*}

We review state-of-the-art embedding methods of MOT on widely used datasets, including MOT17 \cite{milan2016mot16}, MOT20 \cite{dendorfer2020mot20}, and KITTI \cite{geiger2012we}. We include methods published in top CV venues over the past three years. We mainly focus on three evaluation metrics: MOTA, IDF1, and HOTA. The performances on three benchmarks are reported in Table~\ref{tab:SOTA_MOT17}, Table~\ref{tab:SOTA_MOT20}, and Table~\ref{tab:SOTA_KITTI}. The embedding methods in the table include ``Patch-Level Box Image Embedding" (Patch), ``Single-Frame Detection Embedding" (S-Fr), ``Single-Frame Detection Embedding" (X-Fr), ``Correlation-Based Embedding" (Corr), ``Sequential Embedding" (Seq), ``Sequential Embedding" (Tracklet) and ``Sequential Embedding" (X-Track). The used detection can be Public (Pub.) or Private (Private). Modality (Mod.) in the KITTI benchmark includes Vision (V) and LiDAR (L). The Object (Obj.) category includes Car (C) and Person (P). We analyze the top performance for each category of embedding methods as follows.

\subsubsection{General Discussion across Embedding Methods}
We provide a general discussion and comparison across different embedding methods in this section.

\textbf{Detection}.
Since MOT17 and MOT20 provide publicly available detections, early works \cite{yoon2018online,sun2019deep,wang2019exploit} strictly follow the rule of using public detections. Because of the noisy results of public detection, such methods cannot achieve very competitive performance. Then some works \cite{keuper2018motion,bergmann2019tracking,zhou2020tracking} employ refinement techniques that use well-pre-trained models for bounding box regression, significantly alleviating the drawback of noisy public detections. Recently, more works \cite{zhang2021fairmot,wang2021general,yu2022towards} have focused on private detection, which can greatly outperform trackers using public detection.

\textbf{Tracking paradigm}.
Conventional MOT methods \cite{shen2018tracklet,wang2019exploit} follow the tracking-by-detection paradigm. Association is conducted after object detection (usually in bounding box or pose format), where object detection is a separate stage. The tracking performance is largely dependent on the detection performance. 
Since the segmentation mask can provide pixel-wise semantic information, tracking-by-segmentation \cite{voigtlaender2019mots} begins to attract attention after the MOTS dataset is created. The methods achieve superior results compared with the tracking-by-detection counterpart. 
More recently, many works \cite{wang2020towards,zhang2021fairmot} have been interested in end-to-end learning, which follows joint detection and tracking/embedding paradigm. Since the detection error can be corrected simultaneously with embedding learning, promising results are achieved. 
With the fast development of visual Transformers, the tracking-by-attention paradigm \cite{meinhardt2022trackformer} shows leading performance in MOT. 
Moreover, some works \cite{wang2021track} propose a tracking-by-reconstruction scheme that can recover trajectories directly along with the embedding.

\textbf{Model architecture}.
CNN is widely adopted in feature extractors, backbone networks, and Re-ID networks. Commonly used CNNs include VGGNet \cite{sun2019deep}, ResNet \cite{shen2018tracklet,peng2020chained,pang2020tubetk}, and DLA \cite{zhang2021fairmot}. Methods \cite{zhou2020tracking,wang2021joint,zhang2021fairmot,wang2021multiple} show that point-based embedding can better represent object-level features with DLA architecture. 
RNN is commonly used in sequential embeddings. Commonly used architectures include LSTM \cite{lu2017online,chaabane2021deft} and GRU \cite{babaee2019dual,tokmakov2021learning}. RNN alone usually cannot achieve promising results. As a result, embedding methods with RNN also incorporate other architectures like GNN \cite{xu2020train,saleh2021probabilistic} and CNN \cite{guo2021online,tokmakov2021learning} to boost the performance. 
Transformers have gained more attention in recent years. With the global attention mechanism, Transformer-based trackers \cite{xu2021transcenter,chen2022patchtrack,zhao2022tracking} show great potential.
Graph-based architectures are often employed in the head networks for cross-object node embedding \cite{braso2020learning,wang2021joint,zhou2022global}. Promising results are achieved.



\textbf{Computation complexity}. 
Generally speaking, pairwise embedding, including patch pairs and tracklet pairs, increases the computation complexity from $\mathcal{O}(n)$ to $\mathcal{O}(n^2)$, where $n$ is the number of individual objects.
Computation complexity analysis is mainly for measuring the efficiency of the association methods. 
For example, Ullah et al. \cite{ullah2018directed} propose a directed sparse graphical model (DSGM) to reduce computational complexity by minimizing connectivity in the multi-target scene.
Weng et al. \cite{weng20203d} present a baseline of a 3D MOT method with a 3D Kalman filter and the Hungarian algorithm for efficient real-time tracking. 
However, the association methods and computation analysis are well described in existing surveys \cite{park2021multiple}. As a result, the details of this part are not in the main scope of this paper.

\subsubsection{Implementation on Extra Benchmarks}
Since the embedding methods are coupled with other factors like association methods to achieve the final performance, it is better to compare embedding strategies alone and fix other factors the same for evaluation. 

For a fair comparison, we ensure the same training and testing data are used across different embedding methods. Since the ground truth annotation of the testing set in MOT benchmarks is not publicly available, we use a recently released novel MOT benchmark, DIVOTrack \cite{hao2023divotrack}, for reporting the results. 
DIVOTrack is a real-world multi-view multi-person tracking dataset with diverse scenes and dense pedestrians. There are ten scenarios, and each scenario is captured by three cameras. In total, 60 videos are recorded, and each video has about one thousand frames. 

We consider widely used SOTA trackers with released source code for each embedding category in the evaluation, including ByteTrack \cite{zhang2022bytetrack}, FairMOT \cite{zhang2021fairmot}, CenterTrack \cite{zhou2020tracking}, QDTrack \cite{pang2021quasi}, PermaTrack \cite{tokmakov2021learning}, LPC\_MOT \cite{dai2021learning}, and GTR \cite{zhou2022global} from seven different embedding categories, as shown in Table~\ref{tab:SOTA_DIVO}. Note that there are also other trackers without source code available that can provide good performance in MOT benchmarks. However, it is hard to build trackers from scratch with missing implementation details. As a result, we choose widely known SOTA trackers for comparison.

We train the embedding models of selected trackers shown in Table~\ref{tab:SOTA_DIVO} on the MOT17 training set. After training, all the trackers except the LPC\_MOT tracker follow the standard DeepSORT \cite{wojke2017simple} association method with the Hungarian method in the inference to decouple the influence from association strategies. Since the LPC\_MOT tracker is a tracklet-based embedding approach and it is not suitable to use the Hungarian method for the tracklet-based association, we use the default association strategy in LPC\_MOT for reporting the result. As for ByteTrack, we use the YOLOX model \cite{ge2021yolox} for training the detection since it is a patch-based embedding method. The detailed results are shown in Table~\ref{tab:SOTA_DIVO}. LPC\_MOT achieves the best performance in HOTA, IDF1, AssA, IDSW, and Frag, showing the tracklet-based embedding strategy is very effective in the ID association. ByteTrack achieves the second best in HOTA and IDF1, and the best in MOTA, demonstrating that patch-based image embedding can benefit from offline SOTA object detectors like YOLOX. FairMOT and GTR also achieve relatively good performance in HOTA, IDF1, and MOTA, verifying the effectiveness of single-frame embedding and cross-track embedding strategies.

\section{Future Directions of MOT Embedding}

\label{sec:trend}

\subsection{Under-Investigated Areas}

We discuss the trends and potential future directions for MOT embedding that are under-investigated from five different aspects, including non-fully supervised learning, generalization, and domain adaptation, embedding for crowded scenes, multi-view collaboration, and multi-modal MOT.

\subsubsection{Non-Fully Supervised Learning}

Most existing embedding methods in MOT use a fully supervised training framework with object locations and track IDs. However, it is expensive to annotate a large amount of video data from diverse scenarios. Utilizing non-fully annotated video data is in high demand in the MOT area. 

Some works try to adopt weakly supervised and semi-supervised learning frameworks to alleviate the annotation cost. Typically, weakly supervised learning uses the weak label in the annotation for each sample, while semi-supervised learning combines labels and unlabeled samples in training. For example, Ruiz et al. \cite{ruiz2021weakly} employ weakly instance segmentation by Grad-CAM heatmaps \cite{selvaraju2017grad} to extract partial foreground masks for the MOTS task. In place of tracking annotations, McKee et al. \cite{mckee2021multi} first hallucinate videos from images with bounding box annotations using zoom-in/out motion transformations to obtain free tracking labels. Nishimura et al. \cite{nishimura2020weakly} generate pseudo labels for successive frames with the association for cell tracking. Li et al. \cite{li2021semi} adopt contrastive representation learning between detection and sub-tracklet embeddings, combining both labeled tracks and unlabeled tracks in training. 
Some works \cite{karthik2020simple} claim that they propose unsupervised trackers that do not use ground truth tracking labels. However, they assume the box annotation or detection results are available. As a result, these are actually weakly supervised approaches. For example,
Liu et al. \cite{liu2022online} design an unsupervised Re-ID framework based on strong and weak supervision of the association cues from two frames following FairMOT \cite{zhang2021fairmot}.
Wu et al. \cite{wu2020transductive} propose a transductive interactive self-training method to adapt the tracking model to unseen crowded scenes with unlabeled testing data by means of teacher-student interactive learning.
However, these weakly supervised/semi-supervised approaches usually generate pseudo-tracking labels with augmentation and simple trackers that could bring much tracking error. Recently, Yoon et al. \cite{yoon2021weakly} propose a novel masked warping loss that drives the network to indirectly learn how to track objects through a video, trained only with the bounding box information. This type of loss only considers a very short period of temporal consistency. As a result, it cannot achieve competitive performance compared with supervised approaches. 

Some works \cite{bastani2021self,he2019tracking} follow an unsupervised learning framework in MOT. This kind of work takes into consideration of prior consistency and constraints in embedding learning. Specifically, Bastani et al. \cite{bastani2021self} propose a self-supervised tracker by using cross-input consistency, in which two distinct inputs are constructed for the same sequence of video by hiding different information about the sequence in each input. 
He et al. \cite{he2019tracking} propose a tracking-by-animation framework, where a differentiable neural model tracks objects from input frames and then animates these objects into reconstructed frames, which achieves both label-free and end-to-end learning of MOT. 

\subsubsection{Training with Synthetic Data}

Since fully annotated real video data is hard to collect, many works \cite{gaidon2016virtual,liu2021synthetic,fabbri2021motsynth} try to utilize synthetic data for training trackers. The annotations, such as box, mask, and track ID, can be naturally obtained from synthetic data. For example, Gaidon et al. \cite{gaidon2016virtual} show that pre-training on virtual data can improve performance. Liu et al. \cite{liu2021synthetic} introduce a large-scale synthetic data engine named MOTX to generate synthetic data for MOT training and achieve competitive results with the trackers trained using real data. In addition, MOTSynth \cite{fabbri2021motsynth} shows that better performance can be achieved with synthetic data on some widely known trackers such as Tracktor \cite{bergmann2019tracking}, Track R-CNN \cite{voigtlaender2019mots}, Lift\_T \cite{hornakova2020lifted}, MPNTrack \cite{braso2020learning}, and CenterTrack \cite{zhou2020tracking}. These works prove the efficiency of utilizing synthetic data for the MOT task. However, how to reduce the domain gap between synthetic and real data and design a learning framework with the generalization for unseen diverse scenarios is still under-explored.

\subsubsection{Multi-View Collaborated MOT}
In recent years, there have been a few works about multi-view MOT \cite{nguyen2022lmgp,song2020cross,gan2021self,Quach_2021_CVPR,nguyen2021lmgp,xu2016multi,han2021multiple,chen2020cross,han2020complementary,xu2017cross}, where multiple objects are tracked from several different overlapping views simultaneously, aiming to address occlusion issues with multi-view geometry and consistency. The embedding framework is learned for both cross-frames and cross-views. The multi-view also provides depth information, enabling tracking in the 3D space. 
Specifically, 
Xu et al. \cite{xu2016multi} propose a hierarchical composition model and re-formulate multi-view MOT as a problem of compositional structure optimization.
Han et al. \cite{han2020complementary} model the data similarity using appearance, motion, and spatial reasoning and formulate the multi-view MOT as a joint optimization problem solved by constrained integer programming. 
Han et al. \cite{han2021multiple} formulate multi-view MOT as a constrained mixed-integer programming problem and effectively measure the similarity of subjects over time and across views.
Gan et al. \cite{gan2021self} propose a spatial-temporal association network with two designed self-supervised learning losses, including a symmetric-similarity loss and a transitive-similarity loss, to associate multiple humans over time and across views.
In \cite{nguyen2021lmgp}, tracklets are matched to multi-camera trajectories by solving a global lifted multi-cut formulation that incorporates short and long-range temporal interactions on tracklets located in the same camera as well as inter-camera ones.
In addition, Chen et al. \cite{chen2020cross} present a novel solution for multi-human 3D pose estimation and tracking simultaneously from multiple calibrated camera views.

However, there are still unresolved issues for such settings. First, with a limited multi-view collaborated dataset, cross-view Re-ID embedding is not robust and causes degradation in tracking performance. Second, it is difficult to define an end-to-end joint detection and tracking framework that can take multiple views with variant spatial-temporal relations. Third, how to better utilize the cross-frame and cross-view geometry consistency is still not well-explored.

\subsubsection{Multi-Modal MOT}

Except for vision-based trackers, other modalities can also be adopted in MOT. Since LiDAR has a good measure of depth, LiDAR-based trackers \cite{luo2021exploring,pang2021simpletrack,wu2021tracklet,shi2020pv,Sun_2021_ICCV} have become popular in autonomous driving scenarios. 
As different sensors (\textit{e.g}., RGB camera, LiDAR, radar) get
increasingly used together, the multi-modal MOT \cite{linder2016multi,zhang2019robust,zhong2020modeling,chiu2021probabilistic,kim2021eagermot,huang2021joint,nabati2021cftrack,valverde2021there} starts to attract research attention. One benefit is that multiple sensors increase the diversity of object representations, which provides higher association reliability across objects from different timestamps.
Different modalities have their own advantages and disadvantages.
In a tracking-by-detection framework, fusing data from multiple modalities would ideally improve tracking performance than using a single modality.

For example, Linder et al. \cite{linder2016multi} integrate a real-time multi-modal laser/RGB-D people tracking framework for moving platforms.
Zhang et al. \cite{zhang2019robust} propose a generic sensor-agnostic multi-modality MOT framework, where each modality is capable of performing its role independently to preserve reliability, and could further improve its accuracy through a novel multi-modality fusion module.
In \cite{zhong2020modeling}, features of RGB images and 3D LiDAR points are extracted separately before they fully interact with each other through a cross-modal attention mechanism for inter-frame proposals association.
Chiu et al. \cite{chiu2021probabilistic} learn how to fuse features from 2D images and 3D LiDAR point clouds to capture the appearance and geometric information of an object.
Huang and Hao \cite{huang2021joint} present an efficient multi-modal MOT framework with online joint detection and tracking schemes and robust data association for autonomous driving applications. 
CFTrack \cite{nabati2021cftrack} presents an end-to-end network for joint object detection and tracking with a center-based radar-camera fusion algorithm.
Valverde et al. \cite{valverde2021there} propose a novel self-supervised MM-DistillNet framework consisting of multiple teachers that leverage diverse modalities, including RGB, depth, and thermal images, to simultaneously exploit complementary cues and distill knowledge into a single audio student network.

However, there are still unresolved issues for multi-modal MOT. First, the embedding strategy with multi-modality is not well explored, following the simple fusion scheme. Second, except for LiDAR and radar, other modalities are not frequently used. For example, how to combine sound, text, infrared sensors, and other modalities should be explored for some special cases \cite{Scheiner_2020_CVPR,chen2021vehicle}. Third, addressing missing or noisy modalities in MOT is also one of the major challenges.

\subsubsection{MOT with Reinforcement Learning}

In recent years, deep reinforcement learning has gained significant success in various vision applications, such as object detection \cite{liang2017deep}, face recognition \cite{rao2017attention}, and image super-resolution \cite{cao2017attention}. However, deep reinforcement learning remains under-explored in the MOT area, with a limited number of works found \cite{rosello2018multi,jiang2018multiobject,ren2018collaborative,jiang2019multi}. 
For example, in \cite{rosello2018multi}, a novel multi-agent reinforcement learning formulation of MOT is presented, which treats creating, propagating, and terminating object tracks as actions in a sequential decision-making problem. 
Ren et al. \cite{ren2018collaborative} develop a deep prediction-decision network, which simultaneously detects and predicts objects under a unified network via deep reinforcement learning. 

There are several challenges in deep reinforcement learning-based MOT frameworks. First, how defining the space of state and action is crucial. The defined states and actions should be implicit and not annotated with ground truth. Otherwise, they may turn out to become simple supervised learning approaches. Besides that, the defined space should assist the tracking performance. Rather than simple states such as ``tracked" and ``lost", other more complex behaviors can be modeled, such as ``leaving", ``start walking", etc. Second, the definition of reward functions is also important. Reward functions should reflect the relationship between chosen actions and tracking performance based on the defined space. Deep reinforcement learning may gain more potential in the MOT field with carefully designed space and reward functions.

\subsection{Towards Future Embedding Learning in MOT}

\subsubsection{Meta-Learning}

The field of meta-learning \cite{hospedales2020meta,vanschoren2018meta}, or learning-to-learn, has seen a dramatic rise in interest in recent years. Contrary to conventional approaches in machine learning, where tasks are solved from scratch using a fixed learning algorithm, meta-learning aims to improve the learning algorithm itself, given the experience of multiple learning episodes. 
This ``learning-to-learn" \cite{thrun1998learning} can lead to a variety of benefits, such as improved data and compute efficiency, and it is better aligned with human and animal learning \cite{harlow1949formation}, where learning strategies improve both on a lifetime and evolutionary timescales \cite{giraud2004introduction}. Meta-learning has been exploited in many computer vision tasks, such as few-shot object detection \cite{perez2020incremental,kang2019few} and segmentation \cite{shaban2017one,dong2018few,rakelly2018few}. However, meta-learning is still under-explored in the MOT task.

Meta-learning can be a potential novel direction in MOT with the learning-to-learn framework. First, a meta-learner can be adopted to optimize the hyper-parameters, such as the detection and tracking threshold, multi-task loss weighting, and pseudo-label generation. Second, meta-learning can be used for generalization and adaptation purposes to assist MOT in diverse scenarios.

\subsubsection{Learning with Auxiliary Tasks}

Auxiliary learning \cite{liu2019self} is a method to improve the ability of a primary task by training on additional auxiliary tasks alongside the primary task. The sharing of embeddings across tasks results in additional relevant features being available, which otherwise would not have been learned from training only on the primary task. The broader support of these features across new interpretations of input data allows for better generalization of the primary task. Auxiliary learning is similar to multi-task learning \cite{caruana1997multitask}, except that only the performance of the primary task is of importance, and the auxiliary tasks are included purely to assist the primary task.
Applying related learning tasks is one straightforward approach to assist primary tasks.
Toshniwal et al. \cite{toshniwal2017multitask} apply auxiliary supervision with phoneme recognition at intermediate low-level representations to improve the performance of conversational speech recognition. 
In \cite{liebel2018auxiliary}, the chosen auxiliary tasks can be obtained with low efforts, such as global descriptions of a scene, to boost the performance for single scene depth estimation and semantic segmentation.
By carefully choosing a pair of learning tasks, we may also perform auxiliary learning without ground truth labels in an unsupervised manner. 
Jaderberg et al. \cite{jaderberg2016reinforcement} introduce a method for improving agent learning in Atari games by building unsupervised auxiliary tasks to predict the onset of immediate rewards from a short historical context. 
Du et al. \cite{du2018adapting} propose to use cosine similarity as an adaptive task weighting to determine when a defined auxiliary task is useful.
Jin et al. \cite{Jin_2019_CVPR} explore auxiliary training in the context of keypoint embedding representation learning.

To learn embeddings for the MOT task, the defined auxiliary tasks should help improve the discrimination among different objects for the primary task. For example, incorporating unlabeled videos with an unsupervised loss for auxiliary learning, learning pose-based features to improve the discriminative embeddings of objects, and transferring image-based Re-ID features in the embedding models. It would largely broaden the view of researchers with such explorations.

\subsubsection{Large-Scale Pre-Training}

Self-supervised learning has gained popularity because of its ability to avoid the cost of annotating large-scale datasets \cite{jaiswal2021survey}. It can adopt self-defined pseudo labels as supervision and use the learned representations for several downstream tasks. 
With recent self-supervised learning approaches, such as SimCLR \cite{chen2020simple}, SwAV \cite{caron2020unsupervised}, MoCo \cite{he2020momentum}, and video-based tasks \cite{kim2019self,goyal2019scaling,han2020self,qian2021spatiotemporal}, the large-scale pre-training show great power in computer vision tasks.
The pre-training aims at learning the representations with self-supervised pretext tasks with pseudo labels. The learned model from the pretext task can be used for downstream tasks such as classification, detection, and segmentation. 

Since unlabeled video data can be easily obtained, large-scale pre-training can be a good opportunity for learning embeddings for MOT. However, two main issues should be addressed in such a direction. First, defining appropriate pretext tasks to learn object-level embeddings with spatial-temporal information is critical for MOT. Unlike video-level classification tasks, such as anomaly detection \cite{vu2019robust,pang2020self} and action recognition \cite{zhu2018end,chen2021rspnet}, object-level embeddings play an essential role in MOT. Second, how to transfer the pre-training model to the downstream tasks with track IDs is unclear. If these two issues can be addressed, the pre-training approaches will be beneficial to the MOT task with huge unlabeled video data.

\subsubsection{Other Future Directions}
There are other future directions for embedding learning methods in MOT, summarized below. 
\begin{itemize}
\item Distilling knowledge for embedding learning from other tracking-related models, such as image-based Re-ID models and detection models.
\item Learning cross-domain embeddings to bridge the training and testing distribution differences.
\item Mining priors, constraints, and consistencies, such as enter-leave consistency (counting consistency), geometry consistency, and ego-motion consistency.
\item Estimating implicit object behavior status to boost embedding.
\item Reasoning and causality learning for object trajectory estimation.
\end{itemize}

\section{Conclusion}

\label{sec:conclude}

This paper presents a comprehensive survey with an in-depth analysis of embedding methods in multi-object tracking (MOT). We first review the widely used embedding methods in MOT from seven aspects and provide a detailed summary for each embedding category. Then widely used MOT evaluation metrics are summarized. Besides that, an in-depth analysis of state-of-the-art methods is provided. We also point out under-investigated areas and future worth-exploring research directions in the last section, hoping to inspire more thinking of the embedding strategies in MOT.

\section*{Acknowledgement}
This work is supported by National Natural Science Foundation of China (62106219).


%





\ifCLASSOPTIONcaptionsoff
  \newpage
\fi



%




\bibliographystyle{IEEEtran}
\bibliography{IEEEabrv,ref2.bib}

%








\end{document}